\documentclass{article}

\usepackage[final]{neurips_2025}
\usepackage{natbib}
\setcitestyle{authoryear}

\usepackage[utf8]{inputenc} 
\usepackage[T1]{fontenc}    
\usepackage{hyperref}       
\usepackage{url}            
\usepackage{booktabs}       
\usepackage{amsfonts}       
\usepackage{subcaption}     
\usepackage{nicefrac}       
\usepackage{microtype}      
\usepackage{xcolor}         

\definecolor{bluecite}{HTML}{0875b7}
\hypersetup{
  colorlinks=true,
  citecolor=bluecite,
  linkcolor=magenta
}

\usepackage{minitoc}
\usepackage{tocloft}

\usepackage{graphicx}
\usepackage{wrapfig}
\usepackage{caption}
\usepackage{subcaption}
\usepackage{amsmath,amssymb}
\usepackage{array,multirow}
\usepackage{xspace}
\usepackage{tabularx}
\usepackage{float}
\usepackage{siunitx}
\usepackage[capitalise]{cleveref}
\usepackage{blindtext}
\usepackage{algorithm}
\usepackage[noend]{algorithmic}
\newcommand{\FuncSty}[1]{\textnormal{\texttt{#1}}\unskip}
\DeclareMathOperator*{\argmax}{arg\,max}

\newcommand{\poppy}{{\sc poppy }\xspace}
\newcommand{\compass}{{\sc compass}\xspace}
\newcommand{\memento}{{\sc memento}\xspace}
\newcommand{\sgbs}{{\sc sgbs}\xspace}
\newcommand{\cmaes}{{\sc cma-es}\xspace}
\newcommand{\eas}{{\sc eas}\xspace}
\newcommand{\moco}{{\sc moco}\xspace}
\newcommand{\marco}{{\sc marco}\xspace}
\newcommand{\fer}{{\sc fer}\xspace}
\newcommand{\dimes}{{\sc dimes}\xspace}
\newcommand{\pomo}{{\sc pomo}\xspace}

 \newcommand{\del}[1]{}

\title{Memory-Enhanced Neural Solvers \\ for Routing Problems}

\author{
Felix Chalumeau$^{*}$$^1$~~~\textbf{Refiloe Shabe}$^{*}$$^{1}$~~~\textbf{Noah De Nicola}$^{**}$$^2$\\ \textbf{Arnu Pretorius}$^1$~~~~~\textbf{Thomas D. Barrett}$^{\dagger}$$^1$~~~~~\textbf{Nathan Grinsztajn}$^{\dagger}$$^{1}$ \\
\\
$^1$InstaDeep \\
$^2$University of Cape Town
}

\begin{document}

\maketitle
\def\thefootnote{*}\footnotetext{Equal contribution. Corresponding author: \texttt{f.chalumeau@instadeep.com}}\def\thefootnote{\arabic{footnote}}
\def\thefootnote{**}\footnotetext{Work completed during an internship at InstaDeep}\def\thefootnote{\arabic{footnote}}
\def\thefootnote{$\dagger$}\footnotetext{Equal supervision}\def\thefootnote{\arabic{footnote}}

\begin{abstract}

Routing Problems are central to many real-world applications, yet remain challenging due to their (NP-)hard nature. Amongst existing approaches, heuristics often offer the best trade-off between quality and scalability, making them suitable for industrial use. While Reinforcement Learning (RL) offers a flexible framework for designing heuristics, its adoption over handcrafted heuristics remains incomplete. Existing learned methods still lack the ability to adapt to specific instances and fully leverage the available computational budget. Current best methods either rely on a collection of pre-trained policies, or on RL fine-tuning; hence failing to fully utilize newly available information within the constraints of the budget. In response, we present MEMENTO, an approach that leverages memory to improve the search of neural solvers at inference. MEMENTO leverages online data collected across repeated attempts to dynamically adjust the action distribution based on the outcome of previous decisions. We validate its effectiveness on the Traveling Salesman and Capacitated Vehicle Routing problems, demonstrating its superiority over tree-search and policy-gradient fine-tuning; and showing that it can be zero-shot combined with diversity-based solvers. We successfully train all RL auto-regressive solvers on large instances, and verify MEMENTO's scalability and data-efficiency: pushing the state-of-the-art on \num{11} out of \num{12} evaluated tasks.

\begin{figure}[ht]
    \centering \includegraphics[clip, trim=0.5cm 20.5cm 0.5cm 0.5cm, width=0.9\textwidth]{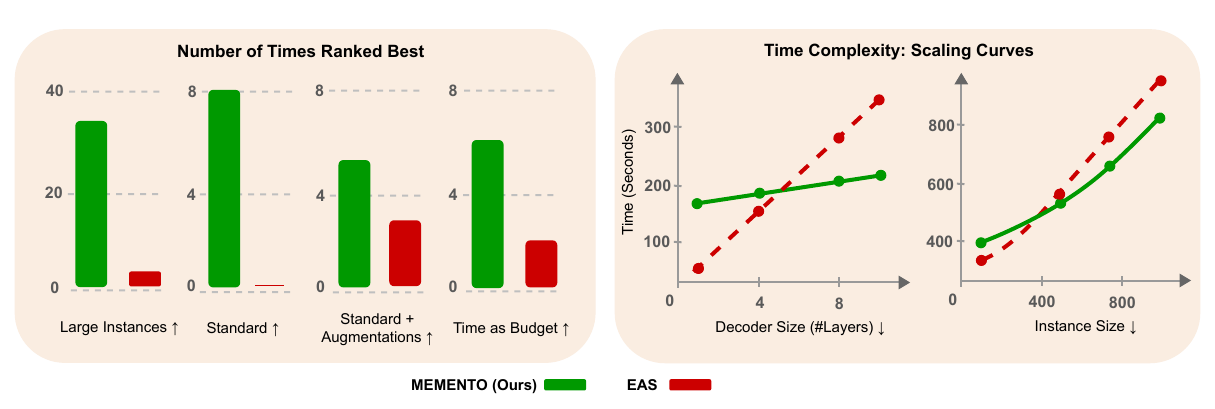}
    \caption{\textbf{MEMENTO outperforms Efficient Active Search} on a wide range of instance sizes and different evaluation budgets. Its time complexity makes it scalable to large instances and policies.}
    \label{fig:moneyshot}
    \vspace{-1.\baselineskip}
\end{figure}

\end{abstract}
\section{Introduction}

Combinatorial Optimization (CO) encompasses a vast range of applications, ranging from transportation~\citep{transportation2011audy} and logistics~\citep{DINCBAS199295} to energy management~\citep{FROGER2016695}. These problems involve finding optimal orderings or labels to optimize given objective functions. Real-world CO problems are typically NP-hard with a solution space growing exponentially with the problem size, making it intractable to find the optimal solution. Hence, industrial solvers rely on sophisticated heuristic approaches to solve them in practice. Reinforcement Learning (RL) provides a versatile framework for learning such heuristics and has demonstrated remarkable success in tackling CO tasks~\citep{MAZYAVKINA2021105400}.

Traditionally, RL methods train policies to incrementally construct solutions. However, achieving optimality in a single attempt for NP-hard problems is impractical. Therefore, pre-trained policies are often combined with search procedures. The literature introduces a range of these procedures, from stochastic sampling~\citep{Kool2019, POMO, poppy}, beam search~\citep{steinbiss1994improvements, choo2022simulationguided}, Monte Carlo Tree Search (MCTS)~\citep{browne2012survey} to online fine-tuning~\citep{Bello16, hottung2022efficient} and searching a space of diverse pre-trained policies~\citep{chalumeau2023combinatorial}. One popular online fine-tuning strategy, Efficient Active Search (\eas) \citep{hottung2022efficient}, uses transitions from generated solutions to derive policy gradient updates. However, it suffers from the inherent drawbacks of back-propagation, in particular having each data point only impacting the update as much as what is enabled by the learning rate. The leading strategy, \compass~\citep{chalumeau2023combinatorial}, relies on a space of diverse pre-trained policies for fast adaptation, but is limited to selecting the most appropriate one, and lacks an update mechanism using the collected data. Meanwhile, the use of memory has grown in modern deep learning, demonstrated by the success of retrieval augmented approaches in Natural Language Processing~\citep{lewis2020retrieval}, and to a smaller extent, its use in RL~\citep{humphreys2022largescale}. These mechanisms create a closer link between collected experience and policy update, making them a promising candidate to improve adaptation.

In this vein, we introduce \memento, a method to update the action distribution of neural solvers online, leveraging a memory of recently collected data. The approach is model agnostic and can be applied to existing neural solvers. \memento learns expressive update rules, which experimentally prove to outperform standard policy-gradient updates. We evaluate our method on two popular routing tasks, the Traveling Salesman Problem (TSP) and the Capacitated Vehicle Routing Problem (CVRP). We evaluate all methods both in and out-of-distribution and tackle instances up to size 500 to better understand their scaling properties. We provide an analysis of the update rules learned by \memento, examining how it enables to adapt faster than policy gradient methods like \eas.

Our contributions come as follows:
\textbf{(i)} We introduce \memento, a memory and a processing module, enabling efficient adaptation of policies at inference time.
\textbf{(ii)} We provide experimental evidence that \memento can be combined with existing approaches to boost performance in and out-of-distribution, even for large instances and unseen solvers, reaching state-of-the-art (SOTA) on \num{11} out of \num{12} tasks.
\textbf{(iii)} Whilst doing so, we train and evaluate leading construction methods on TSP and CVRP instances of size \num{500} solely with RL, outperforming all existing RL methods.
\textbf{(iv)} We open-source our implementations in JAX~\citep{jax2018}, along with test sets and checkpoints.
\section{Related work}
\label{related_work}

\paragraph{Construction methods for CO} These refer to methods that incrementally build a solution one action after the other. \citet{Hopfield85} was the first to use a neural network to solve TSP, followed by~\citet{Bello16} and~\citet{Deudon2018} who respectively added an RL loss and an attention-based encoder. These works were further extended by~\citet{Kool2019} and~\citet{POMO} to use a general transformer architecture~\citep{Attention_all}, which has become the standard model choice that we also leverage in this paper. These works have given rise to several variants, improving the architecture \citep{XinSCZ21,luo2023neural} or the loss~\citep{Kim2021, kim2022symnco, drakulic2023bq, sun2024learning}. These improvements are orthogonal to our work and could a priori be combined with our method. Construction approaches are not restricted to routing problems: numerous works have tackled various CO problems, especially on graphs, like Maximum Cut~\citep{Dai17, barrett2020exploratory}, or Job Shop Scheduling Problem (JSSP)~\citep{Zhang2020, park2021learning}.

Construction methods make use of a predetermined compute budget to generate one valid solution for a problem instance, depending on the number of necessary action steps. In practice, it is common to have a greater, fixed compute budget to solve the problem at hand such that several trials can be attempted on the same instance. However, continuously rolling out the same learned policy is inefficient as (i) the generated solutions lack diversity~\citep{poppy} and (ii) the information gathered from previous rollouts is not utilized. How to efficiently leverage this extra budget has drawn some attention recently. We present the two main types of approaches that augment RL construction methods with no problem-specific knowledge (hence excluding \textit{solution improvement methods}).

\paragraph{Diversity-based methods} The first type of approach focuses on improving the diversity of the generated solutions \citep{POMO, poppy, chalumeau2023combinatorial, hottung2024polynet}. \pomo~\citep{POMO} uses different starting points to enable the same policy to generate diverse candidates. \poppy~\citep{poppy} leverages a population of agents with a loss targeted at specialization on sub-distribution of instances. It was extended in \citet{chalumeau2023combinatorial} and \citet{hottung2024polynet}, respectively replacing the population with a continuous latent space (\compass) or a discrete context vector (PolyNet).

\paragraph{Policy improvement at inference time} The second category of methods, which includes \memento, addresses the improvement aspect. These methods are theoretically orthogonal and can be combined with those mentioned earlier. \eas~\citep{hottung2022efficient} and Meta-SAGE~\citep{son2023metasage} employ a parameter-efficient fine-tuning approach on the test instances, whereas \sgbs~\citep{choo2022simulationguided} enhances this strategy by incorporating tree search. While demonstrating good performance, they rely on rigid, handcrafted improvement and search procedures, which may not be optimal for diverse problems and computational budgets.

In contrast, several methods aim to learn these improvement mechanisms.~\citet{macfarlane2022training} train a Graph Neural Network to perform a tree search akin to MCTS. \moco, \fer, and \marco~\citep{dernedde2024moco, jingwen2023fer, garmendia2024marco} learn the policy improvement update. \moco~\citep{dernedde2024moco} introduces a meta-optimizer that learns to calculate flexible parameter updates based on the reinforce gradient, remaining budget, and best solutions discovered so far. \marco~\citep{garmendia2024marco} aggregates graph edges' features and introduce problem-dependent similarity metrics to improve exploration of the solution space. Both approaches enable to learn adaptation strategies but are tied to heatmap-based policies, preventing their applicability to certain CO problems like CVRP. \fer~\citep{jingwen2023fer} learns a rule to update the instance nodes' embedding, whereas \memento directly updates the action logits, hence being architecture-agnostic.

\section{Methods}
\label{sec:method}

\subsection{Preliminaries}

\paragraph{Formulation}

A CO problem can be represented as a Markov Decision Process (MDP) denoted by $M = (S, A, R, T, H)$. This formulation encompasses the state space $S$ with states \(s_i \in S\), the action space $A$ with actions \(a_i \in A\), the reward function \(R(r |s, a) \), the transition function \(T(s_{i+1}|s_i, a_i)\), and the horizon \(H\) indicating the episode duration. Here, the state of a problem instance is portrayed as the sequence of actions taken within that instance, where the subsequent state \(s_{t+1}\) is determined by applying the chosen action \(a_t\) to the current state \(s_t\). An agent is introduced in the MDP to engage with the problem and seek solutions by learning a policy $\pi(a|s)$. The standard objective considered in RL works is the single shot optimisation, i.e. finding a policy that generates a trajectory \(\tau\) which maximises the collected reward:
$\pi^* = \argmax \nolimits_{\pi} \mathbb{E}_{\tau \sim \pi}[R(\tau)]$, where $R(\tau) = \sum_{t=0}^H R(s_t, a_t)$.

The specificity of most practical cases in RL for CO is that the policy is given a number of attempts allowed per instance (budget \(B\)), to find the best possible solution, rather than a single attempt. Consequently, the learning objective should rather be : $\pi^* = \argmax\nolimits_{\pi} \mathbb{E}_{\tau \sim \pi}[\max\nolimits_{i=1, ..., B}R(\tau_i)]$.

\subsection{MEMENTO}
\label{sec:memento}

Adaptation to unseen instances is crucial for neural solvers. Even when evaluated in the distribution they were trained on, neural solvers are not expected to provide the optimal solution on the first shot, due to the NP-hardness of the problems tackled. Making clever use of the available compute budget for efficient online adaptation is key to performance.

\begin{figure}[ht]
    \centering
    \includegraphics[width=0.95\linewidth]{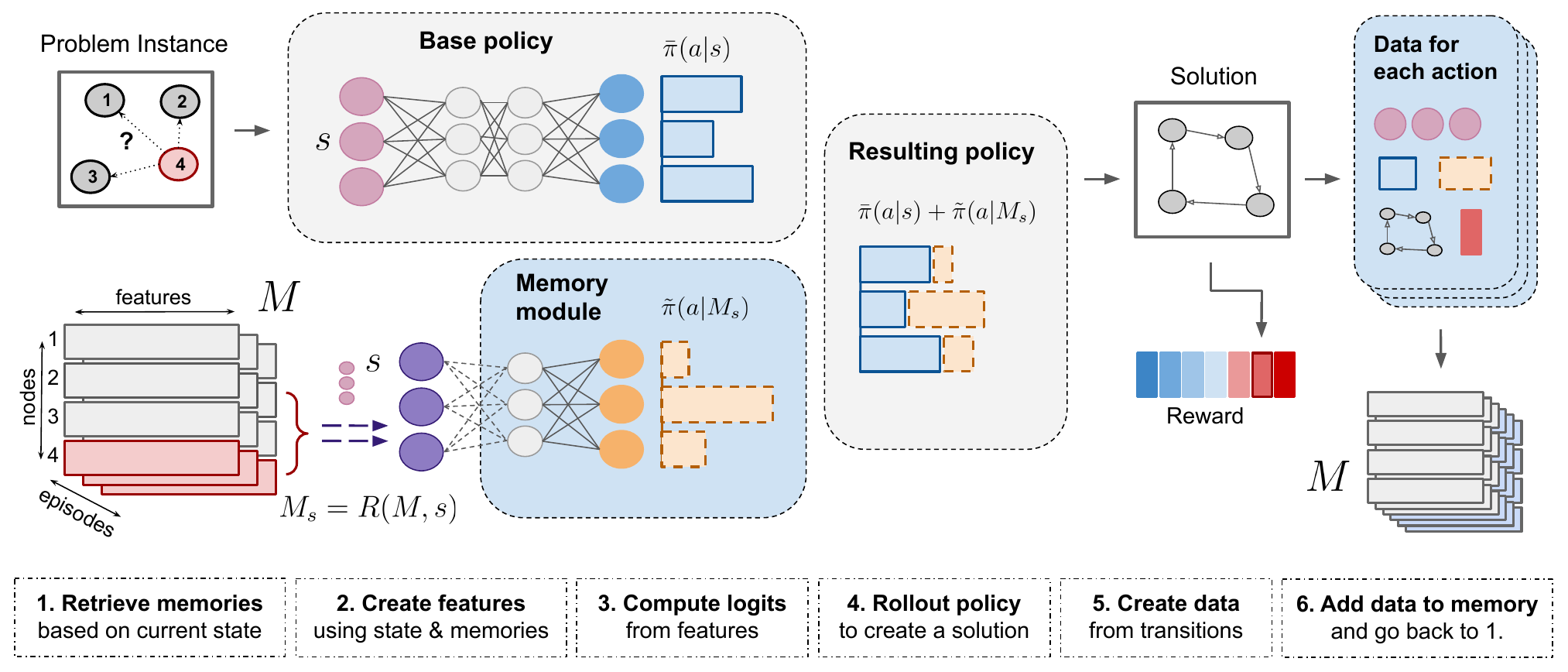}
    \caption{\textbf{MEMENTO uses a memory to adapt neural solvers at inference time.} When taking a decision, data from similar states is retrieved and prepared (1,2), then processed by a MLP to derive correction logits for each action (3). Summing the original and new logits enables to update the action distribution. The resulting policy is then rolled out (4), and transitions' data is stored in a memory (5,6), including node visited, action taken, log probability, and return obtained.}
    \label{fig:memento}
\end{figure}

A compelling approach is to store all past attempts in a memory that can be leveraged for subsequent trajectories. This ensures that no information is lost and that promising trajectories can be used more than once. There are many ways of implementing such framework depending on how the information is stored, retrieved, and used in the policy. We would like the memory-based update mechanism to be (\romannumeral 1) \emph{learnable} (how to use past trajectories to craft better trajectories should be learnt instead of harcoded) (\romannumeral 2) \emph{light-weight} to not compromise inference time unduly (\romannumeral 3) \emph{agnostic} to the underlying model architecture (\romannumeral 4) able to leverage \emph{pre-trained memory-less} policies.

\begin{wrapfigure}{r}{0.54\textwidth}
    \vspace{-1\baselineskip}
    \centering    
    \includegraphics[width=\linewidth]{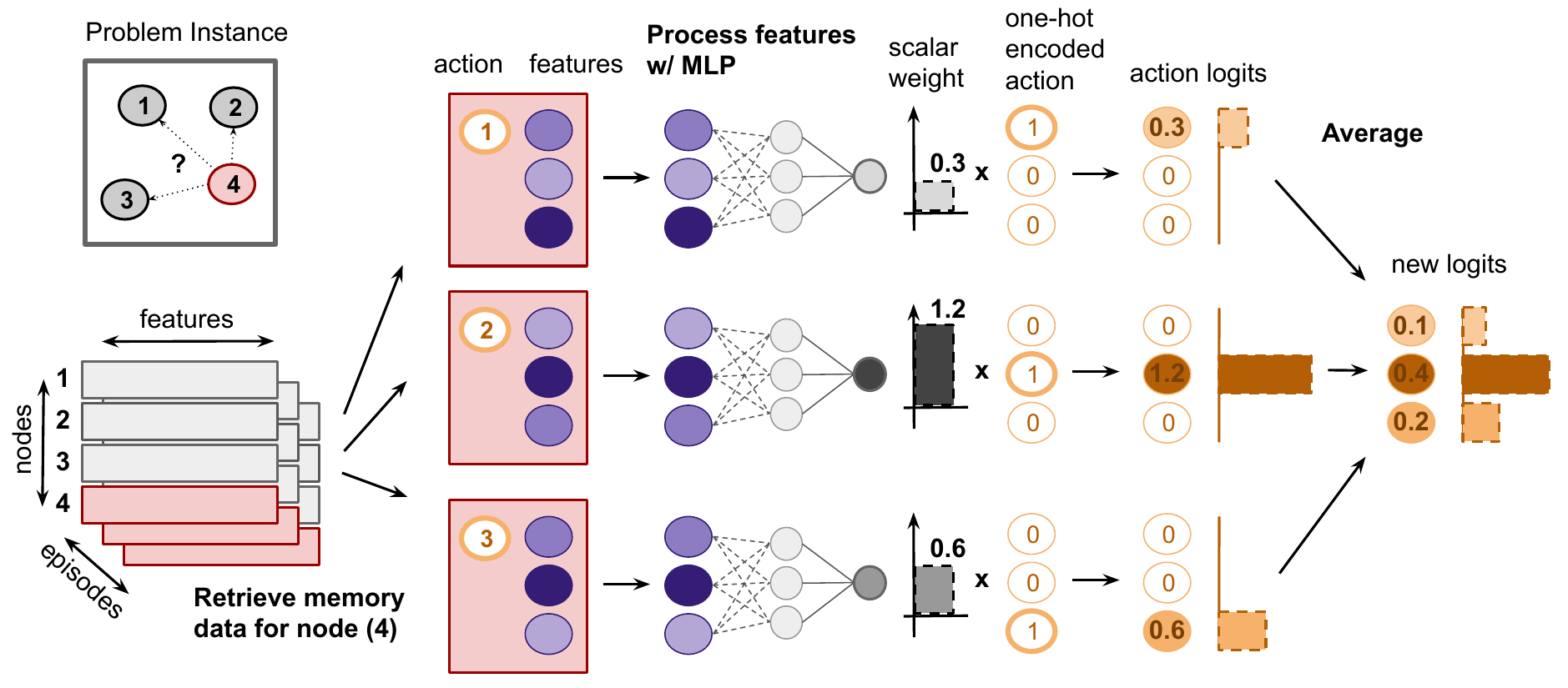}
    \caption{\textbf{Building a new action distribution using the memory.} Relevant data is retrieved and processed by a MLP to derive logits for each possible action.}
    \vspace{-0.4\baselineskip}
    \label{fig:memento2}
\end{wrapfigure}

\paragraph{Overview} To this end, we introduce \memento, a method that dynamically adapts the action distribution of the neural solver using online collected data. This approach, illustrated on~\cref{fig:memento}, achieves the four desiderata: it stores light-weight data about past attempts, and uses an auxiliary model to update the action logits based on the previous outcomes. It can be used on top of any pre-trained policy (architecture-agnostic). In principle, \memento can be combined with existing construction RL algorithms: Attention-Model~\citep{Kool2019}, \pomo~\citep{POMO}, \poppy~\citep{poppy}, \compass~\citep{chalumeau2023combinatorial}, which we confirm experimentally. It learns an update rule during training, enabling data-efficient adaptation compared to policy gradient methods. The data retrieved from the memory and used by the auxiliary model to compute the new action logits contains more than the typical information used to derive a policy gradient update. For instance, the \emph{budget remaining}, enabling to calibrate the exploration/exploitation trade-off and discover superior update rules. We show in~\cref{app:math-reinforce} that \memento has capacity to at least rediscover REINFORCE, and we show empirically in~\cref{sec:memento-eas} that we outperform the leading policy-gradient updates adaptation method.

\paragraph{Storing data in memory} In the memory, we store data about past attempts (\cref{fig:memento}, step 6). Akin to a replay buffer, this data needs to reflect past decisions, their outcome, and must enable to take better future decisions. For each transition experienced while constructing a solution, the memory stores the following: (i) node visited (i.e., in the context of TSP or CVRP, this corresponds to the current city or customer location), (ii) action taken, corresponding to the node that the policy decided to visit next (iii) log-probability given to that action by the model (iv) return of the entire trajectory (negative cost of the solution built) (v) budget at the time that solution was built (vi) log-probability that was suggested by the memory for that action (vii) log-probability associated with the entire trajectory and (viii) the log-probability associated with the remaining part of the trajectory. We can observe that elements (ii, iii and iv) are those needed to reproduce a REINFORCE update, and other features are additional context that will help for credit assignment and distribution shift estimation. The remaining budget is also added to the data when deriving the policy update.

\paragraph{Retrieving data from the memory} Each time an action is taken in a given state, we retrieve data from the memory as shown in step 1 of \cref{fig:memento}. We retrieve data that inform the policy on past decisions that were taken in similar situations. To achieve a good balance between speed and relevance, we retrieve data collected in the same node as we are currently in; showing to be significantly faster than k-nearest neighbour retrieval, while extracting data of similar relevance. More details and motivation for this design choice provided in~\cref{app:memory-retrieval}.

\paragraph{Processing data to update actions} Once data has been retrieved, it is used to derive a policy update. We separate the actions from their associated features (logp, return, etc...). We concatenate the remaining budget to the features. Each feature is normalised, and the resulting feature vector (\cref{fig:memento}, step 2) is processed by a Multilayer Perceptron (MLP) $H_{\theta_M}$ which outputs a \emph{scalar weight}. We compute the logits from the MLP output (\cref{fig:memento}, step 3): each action is one-hot encoded, and a weighted average of the actions vectors is computed based on the scalar weight obtained. This aggregation outputs a new vector of action logits. We sum the new vector of logits with the vector of logits output by the base model. The following paragraph introduces the mathematical formalism, and~\cref{fig:memento2} illustrates it schematically.

More formally, when visiting a node, with a given state $s$, we retrieve from the memory data experienced in the past when visiting that same node. This data, $M_s$, is a sequence of tuples $(a_i, f_{a_i})$, where $a_i$ are the past actions tried and $f_{a_i}$ are various features associated with the corresponding trajectories, as discussed above. The update is computed by $l_M  = \sum_i \mathbf{a_i} H_{\theta_M}(f_{a_i})$, where $\mathbf{a_i}$ is the one-hot encoding of action $a_i$. Let $l$ be the logits from the base policy, the final logits used to sample the next action are given by $l + l_M$. We show in~\cref{app:math-reinforce} that \memento has enough capacity to re-discover the REINFORCE update.

\paragraph{Training}
Existing construction methods are trained for one-shot optimization, with a few exceptions trained for few-shots~\citep{poppy,chalumeau2023combinatorial,hottung2024polynet}. We adapt the training process to fit the multi-shot setting in which methods are used in practice. We rollout the policy (\cref{fig:memento}, step 4-5) as many times as the budget allows on the same instance before taking an update. Our loss is inspired by ECO-DQN~\citep{barrett2020exploratory} and ECORD~\citep{barrett2022ecord}: for a new given trajectory, the reward is the ReLU of the difference between the return and the best return achieved so far. Hence, those summed rewards equal the best score found over the budget. In practice, to account for the fact that it is harder to get an improvement as we get closer to the end of the budget, we multiply this loss with a weight that increases logarithmically as the budget is consumed. See~\cref{appendix:training_process} for explicit formula and pseudo-code.

\paragraph{Inference} The memory processing module can be used with any neural solver. The parameters of the base neural solver and those of the memory processing module are frozen (no more backpropagation), and the action updates are derived by filling the memory with the collected attempts and by processing the data retrieved from the memory, as described in the previous paragraphs, until the inference budget is consumed. All hyperparameters can be found in~\cref{app:hyperparameters}.
\section{Experiments}
\label{experiments}

We evaluate our method across widely recognized routing problems TSP and CVRP. These problems serve as standard benchmarks for evaluating RL-based CO methods~\citep{Deudon2018, Kool2019, POMO, poppy, hottung2022efficient, choo2022simulationguided, chalumeau2023combinatorial}. In~\cref{sec:memento-eas}, we validate our method by comparing it to leading approach for single-policy adaptation using online collected data, \eas~\citep{hottung2022efficient}. We present their comparative performance in line with established standard benchmarking from literature, and elucidate the mechanisms employed to adapt the action distribution of the base policy during inference. We then demonstrate how \memento can be combined with SOTA neural solver \compass with no additional retraining in~\cref{sec:combining-memento}. Finally, we scale those methods to larger instances in~\cref{sec:scaling-rl-construction}, outperforming heatmap-based methods.

\begin{table}[ht]
 \centering
 \caption{\textbf{MEMENTO against baselines on a standard benchmark comprising 8 datasets with 4 distinct instance sizes on (a) TSP and (b) CVRP.} Methods are evaluated on instances from the training distribution (\num{100}) and on larger instance sizes to test generalization. \memento outperforms policy-gradient method \eas and tree-search \sgbs, with significant improvement over the base policy \pomo. \sgbs* results are reported from~\citet{choo2022simulationguided} (bias discussed in~\cref{subsec:std_benchmark}).}
 \label{tab:std_benchmark_summary}
   \begin{subtable}[t]{\textwidth}
    \centering
     \caption{TSP}
      \label{tab:std tsp}
        \scalebox{0.75}{
        \begin{tabular}{l | cc | cc | cc | cc |}
            & \multicolumn{2}{c|}{\textbf{Training distr.}}
            & \multicolumn{6}{c|}{\textbf{Generalization}} \\
          & \multicolumn{2}{c|}{$n=100$} & \multicolumn{2}{c|}{$n=125$} & \multicolumn{2}{c|}{$n=150$} & \multicolumn{2}{c|}{$n=200$} \\
        Method & Obj. & Gap & Obj. & Gap & Obj. & Gap & Obj. & Gap \\
        \midrule
        LKH3 & 7.765 & $0.000\%$ & 8.583 & $0.000\%$ & 9.346 & $0.000\%$ & 10.687 & $0.000\%$ \\
        \midrule
        \pomo (greedy) & 7.796 & 0.404\% & 8.635 & 0.607\% & 9.440 & 1.001\% & 10.933 & 2.300\% \\
        \pomo (sampling) & 7.779 & 0.185\% & 8.609 &  0.299\% & 9.401 & 0.585\% & 10.956 & 2.513\% \\
        \sgbs* & \textit{7.769}* & \textit{$0.058\%$} & - & - & \textit{9.367}* & \textit{$0.220\%$} & \textit{10.753}* & \textit{0.619\%} \\
        \eas-Emb & 7.778 & 0.161\% & 8.604 & 0.238\% & 9.380 & 0.363\% & 10.759 & 0.672\% \\
        \memento & \textbf{7.768} & \textbf{0.045\%} & \textbf{8.592} & \textbf{0.109\%} & \textbf{9.365} & \textbf{0.202\%} & \textbf{10.758} & \textbf{0.664\%} \\
        \end{tabular}}
    \end{subtable}
  \begin{subtable}[t]{\textwidth}
   \centering
    \caption{CVRP}
     \label{tab:std cvrp}
       \scalebox{0.75}{
        \begin{tabular}{l | cc | cc | cc | cc |}
            & \multicolumn{2}{c|}{\textbf{Training distr.}}
            & \multicolumn{6}{c|}{\textbf{Generalization}} \\
          & \multicolumn{2}{c|}{$n=100$} & \multicolumn{2}{c|}{$n=125$} & \multicolumn{2}{c|}{$n=150$} & \multicolumn{2}{c|}{$n=200$} \\
        Method & Obj. & Gap & Obj. & Gap & Obj. & Gap & Obj. & Gap \\
        
        \midrule
        LKH3 & 15.65 & $0.000\%$ & 17.50 & $0.000\%$ & 19.22 & $0.000\%$ & 22.00 & $0.000\%$ \\
        \midrule
        \pomo (greedy) & 15.874 & 1.430\% & 17.818 & 1.818\% & 19.750 & 2.757\% & 23.318 &  5.992\% \\
        \pomo (sampling) & 15.713 & 0.399\% & 17.612 & 0.642\% & 19.488 & 1.393\% & 23.378 & 6.264\% \\
        \sgbs* & \textit{15.659}* & \textit{$0.08\%$} & - & - & \textit{19.426}* & \textit{$1.08\%$} & \textit{22.567}* & \textit{$2.59\%$} \\
        \eas-Emb & 15.663 & 0.081\% & 17.536 & 0.146\% & 19.321 & 0.528\% & 22.541 & 2.460\% \\
        \memento & \textbf{15.657} & \textbf{0.066\%} & \textbf{17.521} & \textbf{0.095\%} &\textbf{19.317} & \textbf{0.478\%} & \textbf{22.492} & \textbf{2.205\%} \\
        \end{tabular}}
\end{subtable}
\vspace{-0.2\baselineskip}
\end{table}

\paragraph{Code availability} We provide access to the code\footnote{Code, checkpoints, and evaluation sets available at \url{https://github.com/instadeepai/memento}} utilized for training our method and executing all baseline models. We release our checkpoints for all problem types and scales, accompanied by the necessary datasets to replicate our findings. We implement our method and experiments in JAX~\citep{jax2018}. The two problems are also JAX implementations from Jumanji~\citep{jumanji2023github}. We use TPU v3-8 for our experiments.

\subsection{Benchmarking MEMENTO against policy gradient fine-tuning}
\label{sec:memento-eas}

The most direct way to leverage online data for policy update is RL/policy-gradients. By contrast, \memento learns an update based on the collected data. We therefore want to see how these two approaches compare. Since \eas~\citep{hottung2022efficient} is SOTA method using policy gradient, we benchmark \memento against it on a standard set of instances used in the literature~\citep{Kool2019, POMO, hottung2022efficient}. Specifically, for TSP and CVRP, these datasets comprise \num{10000} instances drawn from the training distribution. These instances feature the positions of \num{100} cities/customers uniformly sampled within the unit square. The benchmark also includes three datasets of distributions not encountered during training, each comprising \num{1000} problem instances with larger sizes: \num{125}, \num{150}, and \num{200}, generated from a uniform distribution across the unit square. We employ the exact same datasets as those utilized in the literature.

\paragraph{Setup} For routing problems such as TSP and CVRP, \pomo~\citep{POMO} is the base single-agent, one-shot architecture that underpins most RL construction solvers. In this set of experiments, \memento and \eas both use \pomo as a base policy, and adapt it to get the best possible performance within a given budget. We train \memento until it converges, on the same instance distribution as that used for the initial checkpoint. When assessing active-search performance, each method operates within a fixed budget of \num{1600} attempts, a methodology akin to~\citet{hottung2022efficient}. In this setup, each attempt comprises of one trajectory per possible starting point. This standardized approach facilitates direct comparison to \pomo and \eas, which also utilize rollouts from all starting points at each step. We do not use \emph{augmentations with symmetries} in our main results since they consume 87.5\% of the budget to exploit network variance, which blurs the efficacy of the comparison and creates impractical settings for larger instances. Nevertheless, we report them in~\cref{subsec:std_benchmark} for continuity with previous literature.

\paragraph{Results} The average performance of each method across all problem settings are presented in~\cref{tab:std_benchmark_summary}. The observations we draw are three-fold. First, \memento outperforms its base model (\pomo) on the entire benchmark by a significant margin: showing that its adaptive search is superior to stochastic sampling. Second, \memento outperforms \eas on all \num{8} tasks (spanning both in- and out-of-distribution) for both environments, highlighting the efficacy of learned policy updates when compared to vanilla policy gradients. Finally, we note that this improvement is significant; for example, on TSP\num{100}, \memento is doing \num{60}\% better than sampling, while \eas only \num{6}\%. We provide all time costs and performance comparison based on a time budget in~\cref{subsec:std_benchmark}.

\begin{figure}[ht]
    \centering
    \includegraphics[width=0.95\linewidth]{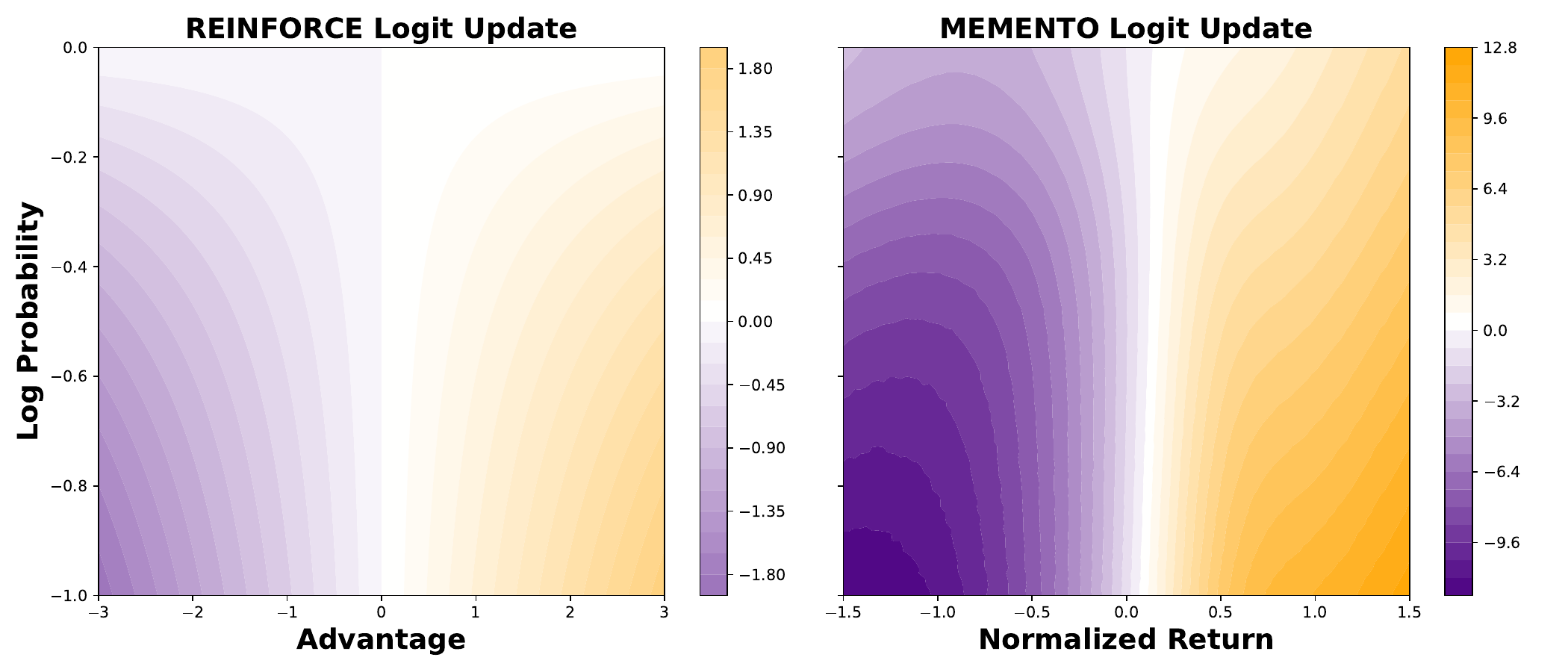}
    \caption{\textbf{Akin to REINFORCE (left), MEMENTO (right) encourages actions with high returns, particularly when they have low probability.} \memento learns an asymmetric rule: requiring the normalised return to be strictly positive to reinforce an action, but encouraging it even more.}
    \label{fig:meta-learned-rules}
\end{figure}

\paragraph{Analysing the update rule} To understand how \memento uses its memory to derive a policy update, we analyse the logit update of an action with respect to its associated data, inspired by~\citet{lu2022discovered, jackson2024discovering}, and compare it to the REINFORCE policy gradient. On~\cref{fig:meta-learned-rules}, we plot the heatmap of the logit update with respect to the log probability (logp) and return associated with an action. We observe that the main rules learned by \memento are similar to REINFORCE: an action with low logp and high return gets a positive update while an action with high logp and low return gets discouraged. 
It is interesting to see discrepancies. In particular, \memento's rules are not symmetric with respect to \(x=0\), it only encourages action that are strictly above the mean return. A similar dissymetry would be expected with \eas since it combines REINFORCE with a term that increases the likelihood to generate the best solution found. \cref{app:update_rule} extends this analysis by visualising the evolution of the update rule for different remaining budget.

Additionally, \memento uses \emph{more inputs than REINFORCE}: current budget available, trajectory logp, age of the data, and previous \memento logit are also used to derive the new action logit. \cref{app:ablation} provides an ablation study of these additional inputs, validating their importance.

\subsection{Zero-shot combination with unseen solvers}
\label{sec:combining-memento}

In this section, we demonstrate zero-shot combination (no retraining) with \compass. \compass adapts by searching amongst a collection of diverse pre-trained policies.

Although providing fast adaptation, it can hardly improve further once the appropriate pre-trained policy has been found. In particular, online collected data cannot be used to update that policy's action distribution. We use \memento's memory module, trained with \pomo as a base policy, apply it on \compass without additional retraining, and show empirically that we can \emph{switch on} \memento's adaptation mechanism during the search and reach a new SOTA on \num{11} out of \num{12} tasks.

\begin{wrapfigure}{r}{0.54\textwidth}
    \vspace{-1.2\baselineskip}
    \centering    
    \includegraphics[width=\linewidth]{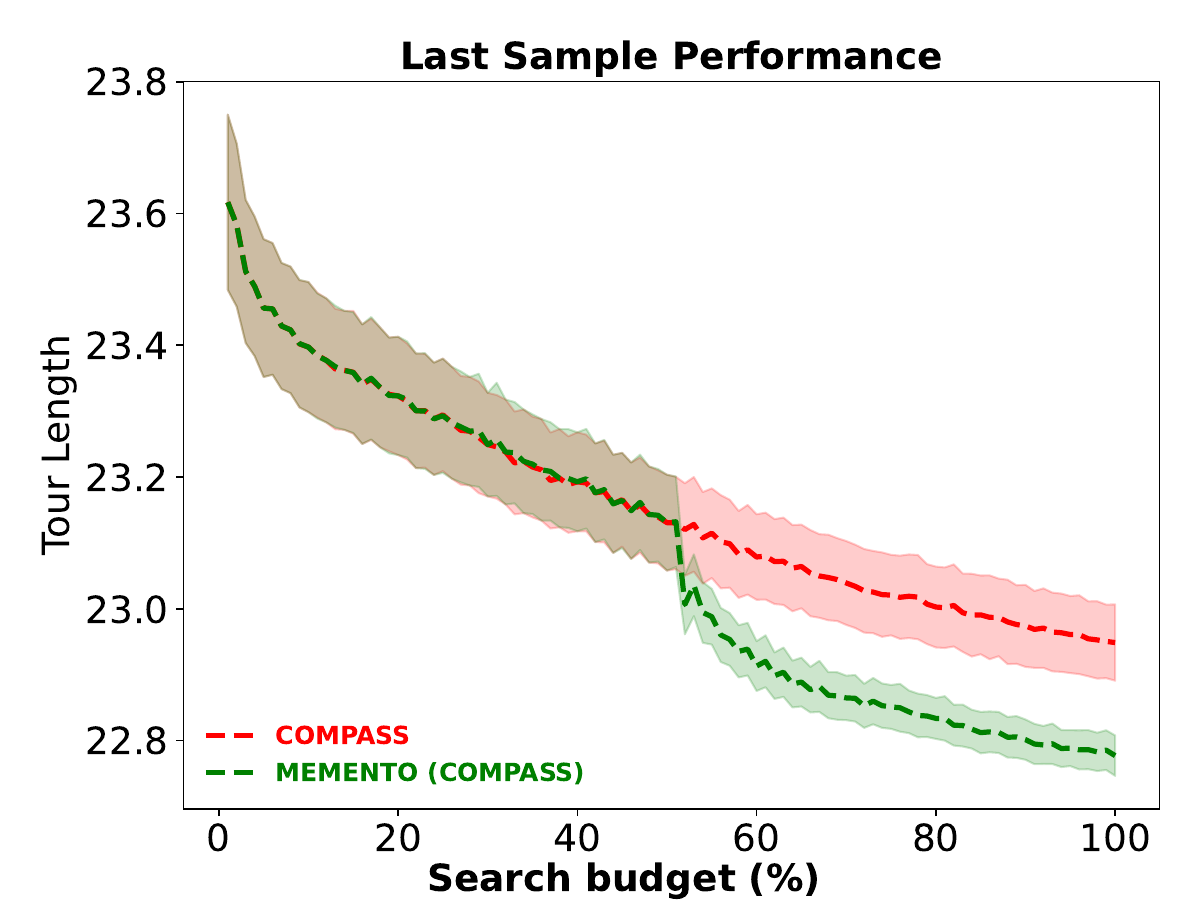}
    \caption{\textbf{Combining MEMENTO and COMPASS} during search on CVRP200, no re-training needed.}
    \vspace{-0.8\baselineskip}
    \label{fig:combining-memento}
\end{wrapfigure}

\paragraph{Methodology} We combine our checkpoint of \memento with a released model of \compass~\citep{chalumeau2023combinatorial}. For the first half of the budget, we do not use \memento, to let \compass find the appropriate pre-trained policy. Once the search starts narrowing down, we activate \memento to adapt the pre-trained policy, with no need to turn off \compass' search. 
\cref{fig:combining-memento} shows the typical trends observed at inference. We recognise the typical search reported in~\citet{chalumeau2023combinatorial}. At \num{50}\% of budget consumption, \memento is activated and one can observe a clear and significant drop in averaged tour length of the latest sampled solutions. The standard deviation also illustrates how the search is narrowed down. 

\paragraph{Results} We evaluate the zero-shot combination of \memento and \compass on the standard benchmark reported in Section~\ref{sec:memento-eas}. Results for CVRP are presented in~\cref{tab:compento}, showing tour length and optimality gap in-distribution (CVRP100) and out-of-distribution. Although requiring no re-training, \memento's rules combine efficiently with \compass and achieves new SOTA. Results on TSP are reported in~\cref{appendix:additional_results}, and results on larger instances (TSP \& CVRP) in~\cref{sec:scaling-rl-construction}. 

\begin{table}[ht]
  \centering
    \caption{\textbf{SOTA performance on CVRP via zero-shot combination} of \memento and \compass. The rules learned by \memento transfer to the population-based method \compass.}
    \label{tab:compento}
    \begin{subtable}[t]{1.\textwidth}
      \centering
       \scalebox{0.75}{
        \begin{tabular}{l | cc | cc | cc | cc |}
    
          & \multicolumn{2}{c|}{$n=100$} & \multicolumn{2}{c|}{$n=125$} & \multicolumn{2}{c|}{$n=150$} & \multicolumn{2}{c|}{$n=200$} \\
        Method & Obj. & Gap & Obj. & Gap & Obj. & Gap & Obj. & Gap \\
        
        \midrule
        
        \begin{tabular}{@{}ll@{}}
        \compass \\
        \memento (\compass) \\
        \end{tabular} &
    
        \begin{tabular}{@{}c@{}}
         15.644\\
        \textbf{15.634} \\
        \end{tabular} &
        
        \begin{tabular}{@{}c@{}}
         -0.019\% \\
        \textbf{-0.082\%} \\
        \end{tabular} &
    
        \begin{tabular}{@{}c@{}}
        17.511 \\
        \textbf{17.497} \\
        \end{tabular} &
        
        \begin{tabular}{@{}c@{}}
        0.064\% \\
        \textbf{-0.041\%} \\
        \end{tabular} &
    
        \begin{tabular}{@{}c@{}}
        19.313 \\
        \textbf{19.290} \\
        \end{tabular} &
    
        \begin{tabular}{@{}c@{}}
        0.485\% \\
        \textbf{0.336\%} \\
        \end{tabular} &
    
        \begin{tabular}{@{}c@{}}
        22.462 \\
        \textbf{22.405} \\
        \end{tabular} &
        
        \begin{tabular}{@{}c@{}}
        2.098\% \\
        \textbf{1.808\%} \\
        \end{tabular} 
        
        \end{tabular}}
    \end{subtable}
\end{table}

\subsection{Scaling RL construction methods to larger instances}
\label{sec:scaling-rl-construction}

Scaling neural solvers to larger instances is a crucial challenge for the field. Auto-regressive construction-based solvers are promising approaches, requiring limited expert knowledge whilst providing strong performance. But are still hardly ever trained on larger scales with RL. As a result, their performance reported in the literature is usually quite unfair, being evaluated on large scales instances (>\num{500}) although having been trained on \num{100} nodes~\citep{qiu2022dimes}. In this section, we explain how we train construction solvers \pomo and \compass on instances of size \num{500} with RL (on TSP and CVRP, checkpoints released); and use them as new base models to train and validate properties of \memento at that scale.

\paragraph{RL training on larger instances} Training \pomo on instances of size \num{500} involves three main steps. First, inspired by curriculum-based methods, we start from a checkpoint pre-trained on TSP100. Second, we use Efficient Attention~\citep{rabe2022selfattention} to reduce the memory cost of multi-head attention, enabling to rollout problems in parallel despite the $O(n^2)$ memory requirement. Third, we use gradient accumulation to keep good estimates despite the constrained smaller instance batch size. These combined tricks enable to train \pomo till convergence within \num{4} days, and consequently train \compass and \memento. We also build the zero-shot combined \memento(\compass) checkpoint.

\begin{figure}[ht]
    \centering \includegraphics[width=0.95\linewidth]{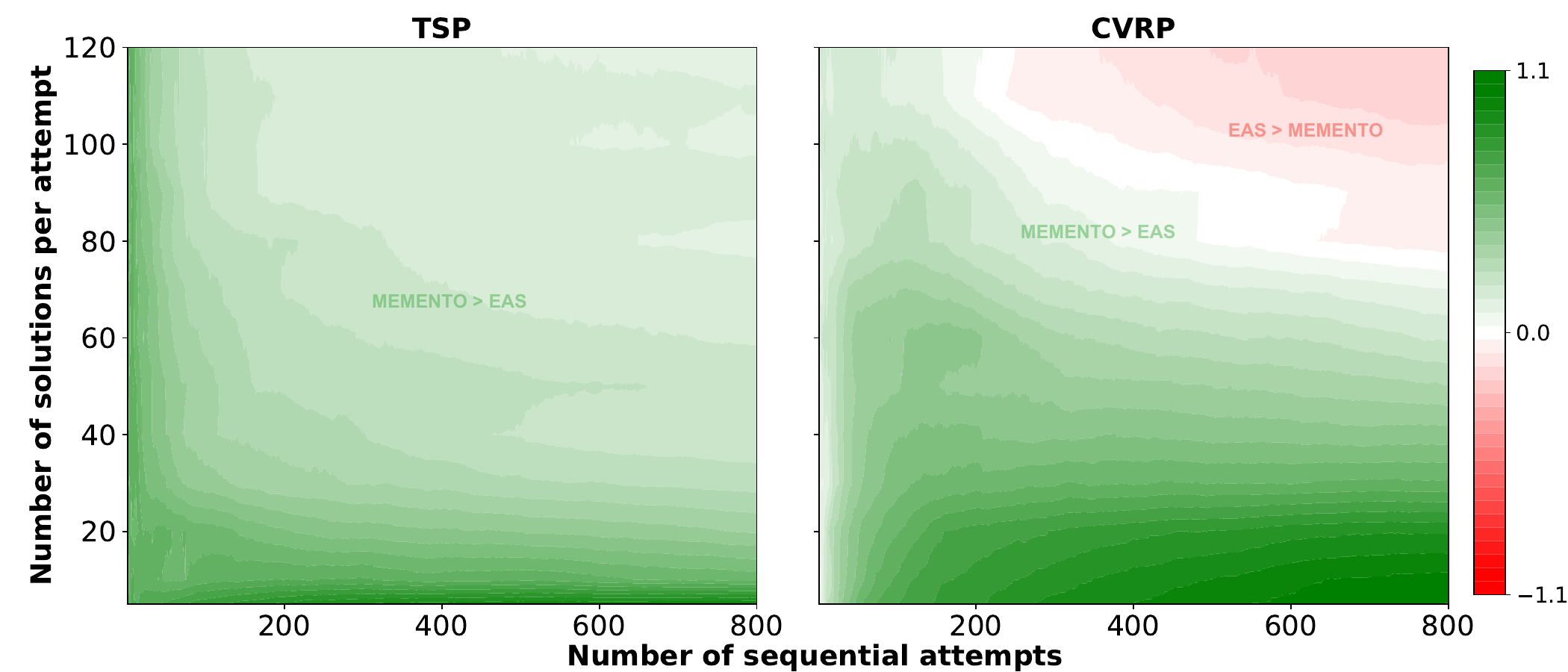}
    \caption{\textbf{MEMENTO outperforms EAS on instances of size 500 across batch sizes and sequential attempts.} Green areas indicate settings where \memento adapts more efficiently. It consistently outperforms \eas on TSP and in most CVRP settings, with strong gains under low budgets.}
    \label{fig:larger-instances}
\end{figure}

\paragraph{Inference-time constraints} On larger instances, the cost of constructing a solution is significantly increased (for \pomo's architecture: squared in computation and memory, and linear in sequential operations). It is also harder to get numerous parallel shots, which most methods depend on. It hence becomes crucial to develop data-efficient methods which stay robust to low budget regimes; and important to know which method to use for each budget constraint.

\paragraph{MEMENTO on larger instances} We compare \memento and \eas on a set of 128 unseen instances (of size \num{500}), for a range of budget expressed in number of sequential attempts and number of parallel attempts attempts. Figure~\ref{fig:larger-instances} reports the percentage of improvement (in absolute cost) brought by \memento over \eas. From this figure, we can observe three main tendencies. First, the data-efficiency of \memento is illustrated by its consistent superiority on low budget (lower left of plots), reaching significant improvement compared to \eas. Then, we can see that, on CVRP, the gap increases with the sequential budget for low parallelism. It demonstrates that \memento's update stays robust, whereas the gradient estimates of \eas get deteriorated and cannot bring further improvement despite additional sequential budget. To finish with, we can see that \memento fully dominates the heatmap on TSP, even for higher budget. However, when budget is increased for CVRP, \eas is able to outperform \memento, given large enough batches ($>80$). 

\begin{wraptable}{r}{0.54\textwidth}
    \vspace{-1.0\baselineskip}
  
  \caption{\textbf{MEMENTO achieves SOTA performance on TSP500.} It outperforms baselines in both budget regimes, attaining single-agent and overall SOTA when combined with \pomo and \compass, respectively.}
  \label{tab:size_500_tsp}
\begin{subtable}[t]{0.54\textwidth}
   \centering
   \scalebox{0.75}{
   \begin{tabular}{l | ccc |}
     Method & Obj. & Gap & Time \\
    \midrule
    Concorde & 16.55 & $0.000\%$ & 38M \\
    LKH-3 & 16.55 & $0.003\%$ & 46M \\
    \midrule
    \dimes & 17.80 & $7.55\%$ & 2H\\
    Difusco & 17.23 & $4.08\%$ & 11M \\
    DeepACO & 18.74 & $13.23\%$ & - \\
    \moco & 16.84 & $1.72\%$ & - \\
    Pointerformer & 17.14 & $3.56\%$ & 1M \\
    \midrule
    \midrule
    & \multicolumn{3}{c|}{\textbf{Low Budget}} \\
    \midrule
    
    \begin{tabular}{@{}ll@{}}
    \pomo (sampling)\\
    \eas \\
    \compass \\
    \memento (\pomo) \\
    \memento (\compass) \\
    \end{tabular} &

    \begin{tabular}{@{}c@{}}
    16.999\\
    16.878\\
    16.811\\
    16.838\\
    \textbf{16.808} \\
    \end{tabular} &
    
    \begin{tabular}{@{}c@{}}
    2.736\%\\
    2.006\%\\
    1.603\%\\
    1.766\%\\
    \textbf{1.586\%}\\
    \end{tabular} &

    \begin{tabular}{@{}c@{}}
    \color{black} 4M \\
    \color{black} 10M \\
    \color{black} 9M \\
    \color{black} 9M \\
    \color{black} 10M 
    \end{tabular} \\
    \midrule
    \midrule

    & \multicolumn{3}{c|}{\textbf{High Budget}} \\
    \midrule
    \begin{tabular}{@{}ll@{}}
    \pomo (sampling)\\
    \eas \\
    \compass \\
    \memento (\pomo) \\
    \memento (\compass) \\
    \end{tabular} 
    
    & \begin{tabular}{@{}c@{}}
    16.986\\
    16.844\\
    16.800\\
    16.823\\
    \textbf{16.795}\\
    \end{tabular} 
    
    & \begin{tabular}{@{}c@{}}
    2.659\%\\
    1.804\%\\
    1.539\%\\
    1.673\%\\
    \textbf{1.507\%}\\
    \end{tabular} 

    & \begin{tabular}{@{}c@{}}
    \color{black} 9M \\
    \color{black} 24M \\
    \color{black} 22M \\
    \color{black} 23M \\
    \color{black} 29M 
    \end{tabular} 
  \end{tabular}}
\end{subtable}
\vspace{-0.8\baselineskip}
\end{wraptable}

\paragraph{Results}
In~\cref{tab:size_500_tsp} and ~\cref{tab:size_500_cvrp}, we report performance of our checkpoints, using datasets introduced in~\citet{Fu_Qiu_Zha_2021} and commonly used in the literature. For each environment, we report the results on two different setting: (i) Low Budget, where the methods are given \num{25000} attempts, (ii) High Budget, where \num{100000} attempts are available. We also report results from concurrent RL methods~\citep{qiu2022dimes, dernedde2024moco, sun2023difusco, ye2023deepaco}, without method-agnostic local search; and industrial solvers LKH3~\citep{lkh3} and Concorde~\citep{concorde}.

First, we observe that stochastically sampling solutions with \pomo for less than \num{10} minutes already provides competitive results, ranking second among the five leading neural solvers. Adding \eas on top of \pomo enables to compete with leading method \moco on TSP500. This shows that auto-regressive RL constructive methods are \emph{competitive at this scale}, contradicting previous literature~\citep{qiu2022dimes,dernedde2024moco}. It is the first time that \pomo is trained at this scale, making it the first RL-trained neural solver that can be used for large instances on both TSP and CVRP: most previous large scale RL methods are graph-specific, and hence cannot be applied on CVRP.

These results confirm that \memento scales to instance size, and can be zero-shot combined with \compass. On this benchmark, \memento achieved SOTA on \num{3} of the \num{4} tasks when zero-shot combined with \compass, showing significant improvement compared to previous SOTA \moco: the gap to optimality is reduced by more than $10\%$. It is the best single agent method on all TSP instances, and on the low budget CVRP task.

\begin{wraptable}{r}{0.54\textwidth}
\vspace{-1.0\baselineskip}
    \caption{\textbf{MEMENTO outperforms baselines on CVRP500.} It achieves single-agent SOTA on low budget regime with \pomo and overall with \compass.}
  \label{tab:size_500_cvrp}
\begin{subtable}[t]{0.54\textwidth}
 \centering
   \scalebox{0.75}{
   \begin{tabular}{l | ccc |}
    Method & Obj. & Gap & Time \\
    \midrule
    LKH-3 & 37.229 & $0.00\%$ & 6H \\
    \midrule
    \midrule
    & \multicolumn{3}{c|}{\textbf{Low Budget}} \\
    \midrule
   \begin{tabular}{@{}ll@{}}
    \pomo (sampling)\\
    \eas \\
    \compass \\
    \memento (\pomo) \\
    \memento (\compass) \\
    \end{tabular} 

    & \begin{tabular}{@{}c@{}}
    37.501\\
    37.425\\
    37.336\\
    37.367\\
    \textbf{37.309}\\
    \end{tabular} 
    
   & \begin{tabular}{@{}c@{}}
    0.731\%\\
    0.525\%\\
    0.287\%\\
    0.369\%\\
    \textbf{0.215\%}\\
    \end{tabular} 

    & \begin{tabular}{@{}c@{}}
    \color{black} 9M \\
    \color{black} 16M \\
    \color{black} 10M \\
    \color{black} 16M \\
    \color{black} 12M 
    \end{tabular} \\
    \midrule
    \midrule
    & \multicolumn{3}{c|}{\textbf{High Budget}} \\
    \midrule
    \begin{tabular}{@{}ll@{}}
    \pomo (sampling)\\
    \eas \\
    \compass \\
    \memento (\pomo) \\
    \memento (\compass) \\
    \end{tabular} 
    
   & \begin{tabular}{@{}c@{}}
    37.456\\
    \textbf{37.185}\\
    37.279\\
    37.306\\
    37.251\\
    \end{tabular} 
    
   & \begin{tabular}{@{}c@{}}
    0.608\%\\
    \textbf{-0.120\%}\\
    0.133\%\\
    0.206\%\\
    0.059\%\\
    \end{tabular} 

    & \begin{tabular}{@{}c@{}}
    \color{black} 20M \\
    \color{black} 36M \\
    \color{black} 25M \\
    \color{black} 65M \\
    \color{black} 36M 
    \end{tabular}
  \end{tabular}}
\end{subtable} 
\vspace{-0.8\baselineskip}
\end{wraptable}

\paragraph{Time complexity} Time performance at scale is key to the adoption of neural solvers, because the number of sequential batches of attempts that can be achieved  within a time budget will mostly depend on it. Search methods must be able to tackle large instances in reasonable time, and must scale well with the size of the base policy used, since~\citet{luo2023neural} demonstrated the importance of using large and multi-layered decoders in neural CO.

Hence, we evaluate \memento and \eas on a set of increasing instance sizes and decoder sizes and report their time complexity on ~\cref{fig:moneyshot} (right). These plots show that, despite being slower for small settings, \memento's adaptation mechanism scales better than \eas. For an instance size of 1000, \memento becomes $20\%$ faster. And even for small instances (size \num{100}), using \num{10} layers in the decoder's architecture (1M parameters) makes \eas $40\%$ slower than \memento. These scaling laws illustrate another benefit of using an approach that learns the update rather than relying on back-propagation to adapt neural solvers at inference time.

\section{Conclusion}
We present \memento, a method to improve adaptation of neural CO solvers to unseen instances by conditioning a policy directly on data collected online during search and search budget. In practice, it proves to outperform stochastic sampling, tree search and policy-gradient fine-tuning, and shows zero-shot combination with an unseen solver. Additionally, \memento respects several key properties, like robustness to low-budget regimes, and favourable time performance scalability. We demonstrate the efficacy of \memento by achieving SOTA single-policy adaptation on a standard benchmark on TSP and CVRP, both in and out-of-distribution. Moreover, we show that \memento finds interpretable update rules to the underlying policy; trading off exploration and exploitation over the search budget and outperforming REINFORCE-style updates. We further demonstrate zero-shot combination of \memento and \compass, achieving overall SOTA on a benchmark of \num{12} tasks. 

\textbf{Limitations and future work} \memento incurs additional compute and memory-usage compared to the memory-less base policies which it augments. Practically, we find that the performance gain significantly outweighs these overheads. A potential mitigation could be to derive the mathematical update learned by \memento to avoid relying on the MLP computation. While we demonstrate successful zero-shot combination of \memento with \compass, training \memento with a diversity-based method or fine-tuning \memento specifically on the \compass policy represents a promising direction for future work.


\section*{Acknowledgements}
We thank the Python and JAX communities for developing tools that made this research possible. We thank the anonymous reviewers for their valuable comments and suggestions that helped improve the final version of this paper. Finally, we thank Google's TPU Research Cloud (TRC) for supporting our research with Cloud TPUs.

\bibliography{references}
\bibliographystyle{abbrvnat}


\newpage

\newpage
\appendix
\section*{Appendix}

\parttoc

\section{Extended results}
\label{appendix:additional_results}

In Section~\ref{experiments}, we compare \memento to popular methods from the literature in numerous settings. In particular, ~\cref{tab:std_benchmark_summary} compares using stochastic sampling, \memento and \eas to adapt \pomo on the standard benchmark; and Figure~\ref{fig:larger-instances} compares the relative performances of \memento and \eas on larger instances of TSP and CVRP for several values of batch sizes and budget. In this section, we report additional values and results of other methods for those experiments.

\subsection{Standard benchmark} 
\label{subsec:std_benchmark}

We evaluate our method, \memento, against leading RL methods and industrial solvers. For RL specific methods, we provide results for \pomo with greedy action selection and stochastic sampling, and \eas; an active search RL method built on top of \pomo, that fine-tunes a policy on each problem instance. We also report population-based method \compass and the zero-shot combination of \compass with \memento.

We compare to the heuristic solver LKH3 \citep{lkh3}; the current leading industrial solver of both TSP and CVRP, as well as an exact solver Concorde \citep{concorde} which is a TSP-specific industrial solver, and the CVRP-specific solver \citep{vidal2012hgs}.

We use datasets of 10,000 instances with 100 cities/customer nodes drawn from the training distribution, and three generalization datasets of 1,000 instances of sizes 125, 150, and 200, all from benchmark sets frequently used in the literature \citep{Kool2019, POMO, hottung2022efficient, poppy, chalumeau2023combinatorial}. Table~\ref{tab:standard_benchmark_tables} displays results for TSP and CVRP on the standard benchmark. The columns provide the absolute tour length, the optimality gap, and the total inference time that each method takes to solve one instance within the attempts budget.

\paragraph{Standard benchmark with augmentation trick}
Augmentation with symmetries is a problem-specific trick that can only be used for a few CO problems. Most prior work assumes that this additional x8 batching can be achieved seamlessly, which is unlikely in practice, when simulating complex real-world scenarios. Using this trick means decreasing the room for search and adaptation, since 87.5\% of the budget is consumed to squeeze performance through uncontrolled network variance, rather than letting methods use principled strategies. Nevertheless, we report the standard benchmark with the "augmentation trick" in Table~\ref{tab:standard_benchmark_aug}. 
It can be observed that \memento outperforms \eas in distribution (CVRP 100), and for the large instances task (CVRP 200). \eas leads on the two other tasks. \memento leads the augmented benchmark overall: it consistently leads in distribution, and both methods are competing closely out-of-distribution. It is to be noted that we have not tuned \memento's hyper-parameters for these runs.

\paragraph{Times reported} When possible, we decide to report the time to solve one instance rather the entire dataset following four main observations: (i) First, the literature use datasets of varying sizes, e.g. 10k for TSP100, 1k for TSP200, 128 for TSP500, hence time reported can be confusing, and do not enable to clearly see how methods' solving time scale with instance size. (ii) Second, the default real-world application consist in solving one instance at a time, or solving several instances at the same time but on separated hardware. (iii) Additionally, measuring on the entire dataset mixes several aspects, i.e. the instance solving time is mixed with the batch scalability of the method. We do think that this property is very interesting to know, but should be considered on the side, rather than mixed with the instance solving time. (iv) Moreover, this property will express differently depending on the hardware available. For instance, on a small hardware \pomo sampling with $n$ attempts will be $n$ times slower than \pomo greedy; but given a large enough GPU, \pomo sampling can be parallelised $n$ times and hence take exactly the same amount of time as \pomo greedy.

We were able to do so on the standard benchmark since we have implementations of most methods we were comparing in a single framework. Since we are reporting several external methods in~\cref{tab:instance_size_500}, we could only report time taken for the full dataset in that case.

\paragraph{Performance with runtime as budget} \label{subsec:runtime_budget}
Expressing budget as time taken to solve one instance is subject to high biases but it is gives interesting perspectives on the time analysis and effects of parallelism. In Table~{\ref{tab:runtime_benchmark}}, we provide results of using runtime in stead of attempts as budget given to each methods to solve TSP and CVRP instances. We use \eas runtime as the reference time.  These results may only be considered moderately since they are subject to a number of limitations: (i) different labs use different implementations and frameworks (ii) labs have access to different hardwares (iii) time is sensitive to parallelism, whereas number of attempts is not. Hence, comparing methods with time is subject to a higher number of biases, and would make it almost impossible to compare papers without a common codebase and hardware. We observe that \memento still leads the benchmark, with 6 out of 8 top results; and both methods are very close on the two tasks where \eas leads.

\paragraph{Additional comments about the results} On~\cref{tab:standard_benchmark_tables}, we can see that in the single-agent setting, \memento leads the entire benchmark, and in the population-based setting, \memento(\compass) is leading the whole benchmark. \memento is able to give significant improvement to \compass on CVRP, pushing significantly the state-of-the-art on this benchmark. On TSP, the improvement is not significant enough to be visible on the rounded results. To finish with, we can observe that the time cost associated with \memento is reasonable and is worth the performance improvement (except maybe for TSP results of \memento(\compass)).

\paragraph{Notes concerning SGBS} All neural methods reported in~\cref{tab:standard_benchmark_tables} are our own runs with standardised checkpoints and rollout strategies, except for the results of \sgbs, which were taken from~\citet{choo2022simulationguided}. This introduces three biases: (i) the base \pomo checkpoint used by \sgbs is not exactly the same as our re-trained \pomo checkpoint (ii) \sgbs uses the domain-specific augmentation trick that we do not use (iii) \sgbs pre-selects starting points in CVRP, which we do not do. 

Overall, the results reported in \sgbs show that \sgbs alone is always outperformed by \eas; hence, it should be expected that in all the settings where we outperform \eas, we would significantly outperform \sgbs if both methods were evaluated exactly in the same way. We hence expect the gap between \memento and \sgbs to be larger than the one reported here. Additionally, \sgbs has yet never been validated on larger instances. Nevertheless, we think that \sgbs is a very efficient method from the NCO toolbox and appreciate that \memento and \sgbs are orthogonal, and could be combined for further improvement. We leave this for future work.

\begin{table}[ht]
  \centering
    \caption{Results of \memento against the baseline algorithms for (a) TSP and (b) CVRP. The methods are evaluated on instances from training distribution ($n = 100$) as well as on larger instance sizes to test generalization. We use the same dataset as the rest of the literature, those contain \num{10000} instances for $n=100$ and \num{1000} instances for $n = 125, 150, 200$. Gaps are computed relative to LKH3. We report time needed to solve one instance. \sgbs* results are reported from~\citet{choo2022simulationguided}, and do not use the same \pomo checkpoint as other reported results. Additionally, they rely on problem-specific tricks that were not used by other methods. Details in~\cref{subsec:std_benchmark}.}
  \label{tab:standard_benchmark_tables}
\begin{subtable}[t]{\textwidth}
\centering
 \caption{TSP}
  \label{tab:standard_tsp}  
    \scalebox{0.62}{
    \begin{tabular}{l | ccc | ccc | ccc | ccc |}
        & \multicolumn{3}{c|}{\textbf{Training distr.}}
        & \multicolumn{9}{c|}{\textbf{Generalization}} \\
      & \multicolumn{3}{c|}{$n=100$} & \multicolumn{3}{c|}{$n=125$} & \multicolumn{3}{c|}{$n=150$} & \multicolumn{3}{c|}{$n=200$} \\
    Method & Obj. & Gap & {\color{black} Time} & Obj. & Gap & {\color{black} Time} & Obj. & Gap & {\color{black} Time} & Obj. & Gap & {\color{black} Time} \\
    \midrule
    Concorde & 7.765 & $0.000\%$ & {\color{black} 0.49S} & 8.583 & $0.000\%$ & {\color{black} 0.72S} & 9.346 & $0.000\%$ & {\color{black} 1S} & 10.687 & $0.000\%$ & {\color{black} 1.86S} \\
    LKH3 & 7.765 & $0.000\%$ & {\color{black} 2.9S} & 8.583 & $0.000\%$ & {\color{black} 4.4S} & 9.346 & $0.000\%$ & {\color{black} 6S} & 10.687 & $0.000\%$ & {\color{black} 11S} \\
    \midrule
    \pomo (greedy) & 7.796 & 0.404\% & \color{black} 0.16S & 8.635 & 0.607\% & \color{black} 0.2S & 9.440 & 1.001\% & \color{black} 0.29S & 10.933 & 2.300\% & \color{black} 0.45S \\
    \pomo (sampling) & 7.779 & 0.185\% & \color{black} 16S & 8.609 &  0.299\% & \color{black} 20S & 9.401 & 0.585\% & \color{black} 29S & 10.956 & 2.513\% & \color{black} 45S \\
    \textit{\sgbs}* & \textit{7.769}* & \textit{0.058\%}* & - & - & - & - & \textit{9.367}* & \textit{0.220\%}* & - & \textit{10.753}* & \textit{0.619\%}* & -\\
    \eas & 7.778 & 0.161\% & \color{black} 39S & 8.604 & 0.238\% & \color{black} 46S & 9.380 & 0.363\% & \color{black} 64S & 10.759 & 0.672\% & \color{black} 91S \\
    \memento (\pomo) & 7.768 & 0.045\% & \color{black} 43S & 8.592 & 0.109\% & \color{black} 52S & 9.365 & 0.202\% & \color{black} 77S & 10.758 & 0.664\% & \color{black} 115S \\
    \midrule
    
    \begin{tabular}{@{}ll@{}}
    \compass \\
    \memento (\compass)
    \end{tabular} &

    \begin{tabular}{@{}c@{}}
    7.765 \\
    \textbf{7.765}
    \end{tabular} &
    
    \begin{tabular}{@{}c@{}}
    0.008 \% \\
    \textbf{0.008\%} 
    \end{tabular} &

    \begin{tabular}{@{}c@{}}
    \color{black}  20S \\
    \color{black} 32S 
    \end{tabular} &

    \begin{tabular}{@{}c@{}}
    8.586 \\
    \textbf{8.586}
    \end{tabular} &
    
    \begin{tabular}{@{}c@{}}
    0.036 \% \\
    \textbf{0.035\%}
    \end{tabular} &

    \begin{tabular}{@{}c@{}}
    \color{black}  24S\\
    \color{black} 39S 
    \end{tabular} &

    \begin{tabular}{@{}c@{}}
    9.354 \\
    \textbf{9.354}
    \end{tabular} &
    
    \begin{tabular}{@{}c@{}}
    0.078\% \\
    \textbf{0.077\%}
    \end{tabular} &

    \begin{tabular}{@{}c@{}}
    \color{black}  33S\\
    \color{black} 58S 
    \end{tabular} &

    \begin{tabular}{@{}c@{}}
    10.724 \\
    \textbf{10.724}
    \end{tabular} &
    
    \begin{tabular}{@{}c@{}}
    0.349\% \\
    \textbf{0.348\%}
    \end{tabular} &

    \begin{tabular}{@{}c@{}}
    \color{black}  49S\\
    \color{black} 88S 
    \end{tabular} 
    \end{tabular}}
\end{subtable}

\begin{subtable}[t]{\textwidth}
\centering
 \caption{CVRP}
  \label{tab:standard_cvrp}
   \scalebox{0.62}{
    \begin{tabular}{l | ccc | ccc | ccc | ccc |}
        & \multicolumn{3}{c|}{\textbf{Training distr.}}
        & \multicolumn{9}{c|}{\textbf{Generalization}} \\
      & \multicolumn{3}{c|}{$n=100$} & \multicolumn{3}{c|}{$n=125$} & \multicolumn{3}{c|}{$n=150$} & \multicolumn{3}{c|}{$n=200$} \\
    Method & Obj. & Gap & {\color{black} Time} & Obj. & Gap & {\color{black} Time} & Obj. & Gap & {\color{black} Time} & Obj. & Gap & {\color{black} Time} \\
    
    \midrule
    HGS & 15.563 & $-0.536\%$ & 19S & - & - & - & 19.055 & $-0.884\%$ & 32S & 21.766 & $-1.096\%$ & 61S \\
    LKH3 & 15.646 & $0.000\%$ & 52S & 17.50 & $0.000\%$ & - & 19.222 & $0.000\%$ & 72S & 22.003 & $0.000\%$ & 90S \\
    \midrule
    \pomo (greedy) & 15.874 & 1.430\% & \color{black} 24S & 17.818 & 1.818\% & \color{black} 34S & 19.750 & 2.757\% & \color{black} 52S & 23.318 &  5.992\% & \color{black} 87S \\
    \pomo (sampling) & 15.713 & 0.399\% & \color{black} 24S & 17.612 & 0.642\% & \color{black} 34S & 19.488 & 1.393\% & \color{black} 52S & 23.378 & 6.264\% & \color{black} 87S \\
    \textit{\sgbs}* & \textit{15.659}* & \textit{0.08\%}* & - & - & - & - & \textit{19.426}* & \textit{1.08\%}* & - & \textit{22.567}* & \textit{2.59\%}* & - \\
    \eas & 15.663 & 0.081\% & \color{black} 66S & 17.536 & 0.146\% & \color{black} 82S & 19.321 & 0.528\% & \color{black} 123S & 22.541 & 2.460\% & \color{black} 179S \\
    \memento (\pomo) & 15.657 & 0.066\% & \color{black} 118S & 17.521 & 0.095\% & \color{black} 150S & 19.317 & 0.478\% & \color{black} 169S & 22.492 & 2.205\% & \color{black} 392S \\
    \midrule
    
    \begin{tabular}{@{}ll@{}}
    \compass \\
    \memento (\compass) \\
    \end{tabular} &

    \begin{tabular}{@{}c@{}}
     15.644\\
    \textbf{15.634} \\
    \end{tabular} &
    
    \begin{tabular}{@{}c@{}}
     -0.019\% \\
    \textbf{-0.082\%} \\
    \end{tabular} &

    \begin{tabular}{@{}c@{}}
     \color{black} 29S\\
    \color{black} 82S 
    \end{tabular} &

    \begin{tabular}{@{}c@{}}
    17.511 \\
    \textbf{17.497} \\
    \end{tabular} &
    
    \begin{tabular}{@{}c@{}}
    0.064\% \\
    \textbf{-0.041\%} \\
    \end{tabular} &

    \begin{tabular}{@{}c@{}}
    \color{black} 39S\\
    \color{black}  100S 
    \end{tabular} &

    \begin{tabular}{@{}c@{}}
    19.313 \\
    \textbf{19.290} \\
    \end{tabular} &

    \begin{tabular}{@{}c@{}}
    0.485\% \\
    \textbf{0.336\%} \\
    \end{tabular} &

    \begin{tabular}{@{}c@{}}
    \color{black} 56S\\
    \color{black} 118S 
    \end{tabular} &

    \begin{tabular}{@{}c@{}}
    22.462 \\
    \textbf{22.405} \\
    \end{tabular} &
    
    \begin{tabular}{@{}c@{}}
    2.098\% \\
    \textbf{1.808\%} \\
    \end{tabular} &

    \begin{tabular}{@{}c@{}}
    \color{black} 85S\\
    \color{black}  272S 
    \end{tabular}
    \end{tabular}}
\end{subtable}
\end{table} 

\begin{table}[ht]
  \centering
    \caption{Results of \memento and the baseline algorithms with instance augmentation for (a) TSP and (b) CVRP.}
  \label{tab:standard_benchmark_aug}
\begin{subtable}[t]{\textwidth}
\centering
 \caption{TSP}
  \label{tab:standard_tsp_aug}
    \scalebox{0.65}{
    \begin{tabular}{l | ccc | ccc | ccc | ccc |}
        & \multicolumn{3}{c|}{\textbf{Training distr.}}
        & \multicolumn{9}{c|}{\textbf{Generalization}} \\
      & \multicolumn{3}{c|}{$n=100$} & \multicolumn{3}{c|}{$n=125$} & \multicolumn{3}{c|}{$n=150$} & \multicolumn{3}{c|}{$n=200$} \\
    Method & Obj. & Gap & { Time} & Obj. & Gap & { Time} & Obj. & Gap & { Time} & Obj. & Gap & { Time} \\
    \midrule
    LKH3 & 7.765 & $0.000\%$ & 2.9S & 8.583 & $0.000\%$ & 4.4S & 9.346 & $0.000\%$ & 6S & 10.687 & $0.000\%$ & 11S \\
    \midrule
    \sgbs & 7.769 & 0.058\% & - & - & - & - & 9.367 & 0.220\% & - & 10.753 & 0.619\% & -\\
    \pomo (sampling) & 7.767 & 0.026\% &  16S & 8.594 &  0.128\% &  20S & 9.376 & 0.321\% &  29S & 10.916 & 2.14\% &  45S \\
    \eas & 7.768 & 0.038\% &  39S & 8.590 & 0.080\% &  46S & 9.361 & 0.159\% &  64S & \textbf{10.730} & \textbf{0.403\%} &  91S \\
    \memento (\pomo) & \textbf{7.765} & \textbf{0.008\%} & 43S & \textbf{8.586} & \textbf{0.035\%} & 52S  & \textbf{9.355} & \textbf{0.091\%} &  77S & 10.743 & 0.526\% & 115S \\
\end{tabular}}
\end{subtable}
\begin{subtable}[t]{\textwidth}
 \centering
 \caption{CVRP}
  \label{tab:standard_cvrp_aug}
   \scalebox{0.65}{
    \begin{tabular}{l | ccc | ccc | ccc | ccc |}
        & \multicolumn{3}{c|}{\textbf{Training distr.}}
        & \multicolumn{9}{c|}{\textbf{Generalization}} \\
      & \multicolumn{3}{c|}{$n=100$} & \multicolumn{3}{c|}{$n=125$} & \multicolumn{3}{c|}{$n=150$} & \multicolumn{3}{c|}{$n=200$} \\
    Method & Obj. & Gap & { Time} & Obj. & Gap & { Time} & Obj. & Gap & { Time} & Obj. & Gap & { Time} \\
    
    \midrule
    HGS & 15.563 & $-0.536\%$ & - & - & - & - & 19.055 & $-0.884\%$ & - & 21.766 & $-1.096\%$ & - \\
    LKH3 & 15.646 & $0.000\%$ & - & 17.50 & $0.000\%$ & - & 19.222 & $0.000\%$ & - & 22.003 & $0.000\%$ & - \\
    \midrule
    \sgbs & 15.659 & 0.08\% & - & - & - & - & 19.426 & 1.08\% & - & 22.567 & 2.59\% & - \\
    \pomo (sampling) & 15.67 & 0.18\% & 24S  & 17.56 & 0.33\% & 34S  & 19.43 & 1.08\% & 52S  & 23.24 & 5.64\% &  87S \\
    \eas & 15.623 & -0.175\% & 66S  & \textbf{17.473} & \textbf{-0.153\%} &  82S & \textbf{19.261} & \textbf{0.213\%} &  123S & 22.556 & 2.49\% & 179S  \\
    \memento (\pomo) & \textbf{15.616} & \textbf{-0.196\%} &  118S &  17.511 & 0.040\% & 150S & 19.316 & 0.477\% &  169S &  \textbf{22.515} & \textbf{2.308\%} &  392S \\
    \end{tabular}}
\end{subtable}
\end{table} 

\begin{table}[ht]
  \centering
    \caption{Results of \memento and the baseline algorithms with budget expressed as runtime for (a) TSP and (b) CVRP.}
  \label{tab:runtime_benchmark}
\begin{subtable}[t]{\textwidth}
\centering
\caption{TSP}
    \scalebox{0.69}{
    \begin{tabular}{l | ccc | ccc | ccc | ccc |}
        & \multicolumn{3}{c|}{\textbf{Training distr.}}
        & \multicolumn{9}{c|}{\textbf{Generalization}} \\
      & \multicolumn{3}{c|}{$n=100$} & \multicolumn{3}{c|}{$n=125$} & \multicolumn{3}{c|}{$n=150$} & \multicolumn{3}{c|}{$n=200$} \\
    Method & Obj. & Gap & { Time} & Obj. & Gap & { Time} & Obj. & Gap & { Time} & Obj. & Gap & { Time} \\
    \midrule
    LKH3 & 7.765 & 0.000\% & - & 8.583 & 0.000\% & - & 9.346 & 0.000\% & - & 10.687 & 0.000\% & - \\
    \midrule
    \pomo (sampling) & 7.779 & 0.216\% &  39S & 8.609 &  0.299\% &  46S & 9.401 & 0.585\% &  64S & 10.956 & 2.513\% &  91S \\
    \eas & 7.778 & 0.161\% &  39S & 8.604 & 0.238\% &  46S & 9.380 & 0.363\% &  64S & \textbf{10.759} & \textbf{0.672\%} &  91S \\
    \memento & \textbf{7.768} & \textbf{0.046\%} & 39S & \textbf{8.592} & \textbf{0.110\%} & 46S  & \textbf{9.365} & \textbf{0.203\%} & 64S  & 10.760 & 0.681\% & 91S  \\ 
    \end{tabular}}
\end{subtable}
\begin{subtable}[t]{\textwidth}
\centering
\caption{CVRP}
   \scalebox{0.69}{
    \begin{tabular}{l | ccc | ccc | ccc | ccc |}
        & \multicolumn{3}{c|}{\textbf{Training distr.}}
        & \multicolumn{9}{c|}{\textbf{Generalization}} \\
      & \multicolumn{3}{c|}{$n=100$} & \multicolumn{3}{c|}{$n=125$} & \multicolumn{3}{c|}{$n=150$} & \multicolumn{3}{c|}{$n=200$} \\
    Method & Obj. & Gap & { Time} & Obj. & Gap & { Time} & Obj. & Gap & { Time} & Obj. & Gap & { Time} \\
    
    \midrule
    LKH3	& 15.647 & 0.000\% & - & 17.504 & 0.000\% & - &	19.225 & 0.000\% &	- & 22.007 & 0.000\% & - \\	
    \midrule
    \pomo (sampling) & 15.713 & 0.399\% &  66S & 17.612 & 0.642\% &  82S & 19.488 & 1.393\% &  123S & 23.378 & 6.264\% &  179S \\
    \eas & 15.663 & 0.081\% &  66S & 17.536 & 0.146\% &  82S & 19.321 & 0.528\% &  123S & \textbf{22.541} & \textbf{2.460\%} &  179S \\
    \memento & \textbf{15.660} & \textbf{0.086\%} & 66S  &  \textbf{17.526} & \textbf{0.127\%} & 82S  & \textbf{19.321} & \textbf{0.502\%} & 123S  &  22.546 & 2.450\% & 179S  \\
    \end{tabular}}
\end{subtable}
\end{table} 

\subsection{Performance over different number of parallel and sequential attempts} \label{subsec:parallel_and_sequential}

In Section \ref{sec:scaling-rl-construction}, we show performance improvement of \memento over \eas on instances of size 500 on TSP and CVRP for increasing number of sequential attempts and size of attempt batch. We report extended results with an additional competitor, \pomo. We compare the online adaptation of the methods over four different sizes of batched solutions across increasing sequential attempts for each CO problem. Results from Table~\ref{tab:instance_size_500} show results of the three methods on TSP and CVRP. The columns show the best tour length performance for various values of sequential attempts expressed as budget, and solution batches of size $N$. Performance is averaged over a set of 128 instances. 

\begin{table}[ht]
  \centering
    \caption{Results of \memento, \pomo and \eas on instances of size 500 of (a) TSP and (b) CVRP for increasing number of sequential attempts and sizes of attempt batch.}
     \label{tab:instance_size_500}
\begin{subtable}[t]{\textwidth}
\centering
 \caption{TSP}
   \label{tab:tsp500}
    \scalebox{0.7}{
    \begin{tabular}{l | ccccc | ccccc |}
        & \multicolumn{5}{c|}{$N = 20$}
        & \multicolumn{5}{c|}{$N = 40$} \\
     Method & 200 & 400 & 600 & 800 & 1000 & 200 & 400 & 600 & 800 & 1000 \\
    \midrule
    \pomo (sampling) & 16.9810 & 16.9707 & 16.9638 & 16.9619 & 16.9574 & 16.9717 & 16.96 & 16.9549 & 16.9523 & 16.9493 \\
    \eas & 16.9223 & 16.8923 & 16.8754 & 16.864 & 16.8556 & 16.8777 & 16.8468 & 16.83 & 16.8208 & 16.8143 \\
    \memento & \textbf{16.8312} & \textbf{16.81} & \textbf{16.799} & \textbf{16.7939} & \textbf{16.7902} & \textbf{16.8187} & \textbf{16.8018} & \textbf{16.7926} & \textbf{16.7855} & \textbf{16.7815} \\
    \midrule
    \midrule
    
    & \multicolumn{5}{c|}{$N = 60$}
    & \multicolumn{5}{c|}{$N = 80$} \\
    Method & 200 & 400 & 600 & 800 & 1000 & 200 & 400 & 600 & 800 & 1000 \\
    \midrule
    
    \begin{tabular}{@{}ll@{}}
    \pomo (sampling)\\
    \eas \\
    \memento 
    \end{tabular} &

    \begin{tabular}{@{}c@{}}
    16.9678 \\
    16.8616 \\
    \textbf{16.8129} 
    \end{tabular} &

    \begin{tabular}{@{}c@{}}
    16.9587 \\
    16.8353 \\
    \textbf{16.7966}
    \end{tabular} &

    \begin{tabular}{@{}c@{}}
    16.9539 \\
    16.8211 \\
    \textbf{16.7867} 
    \end{tabular} &

    \begin{tabular}{@{}c@{}}
    16.952 \\
    16.8134 \\
    \textbf{16.7816} 
    \end{tabular} &

    \begin{tabular}{@{}c@{}}
    16.9503\\
    16.8084\\
    \textbf{16.7780}
    \end{tabular} &

    \begin{tabular}{@{}c@{}}
    16.9634\\
    16.8521\\
    \textbf{16.8087} 
    \end{tabular} &

    \begin{tabular}{@{}c@{}}
    16.9547 \\
    16.8234 \\
    \textbf{16.7935} 
    \end{tabular} &

    \begin{tabular}{@{}c@{}}
    16.9513 \\
    16.8106 \\
    \textbf{16.7854} 
    \end{tabular} &

    \begin{tabular}{@{}c@{}}
    16.9495\\
    16.8014\\
    \textbf{16.7811}
    \end{tabular} &

    \begin{tabular}{@{}c@{}}
    16.9474\\
    16.7941\\
    \textbf{16.7770}
    \end{tabular}
    \end{tabular}}
\end{subtable}
\hfill
\begin{subtable}[t]{\textwidth}
  \centering
   \caption{CVRP}
   \label{tab:cvrp500}
    \scalebox{0.7}{
    \begin{tabular}{l | ccccc | ccccc |}
      & \multicolumn{5}{c|}{$N = 20$}
        & \multicolumn{5}{c|}{$N = 40$} \\
     Method & 200 & 400 & 600 & 800 & 1000 & 200 & 400 & 600 & 800 & 1000 \\
    \midrule
    \pomo (sampling) & 37.5256 & 37.4917 & 37.4763 & 37.4639 & 37.4570 & 37.4858 & 37.4599 & 37.4475 & 37.4372 & 37.4307 \\
    \eas & 37.6107 & 37.5988 & 37.5914 & 37.582 & 37.5769 & 37.5022 & 37.4546 & 37.422 & 37.4017 & 37.3849 \\
    \memento & \textbf{37.3398} & \textbf{37.2992} & \textbf{37.2731} & \textbf{37.2589} & \textbf{37.2527} & \textbf{37.3172} & \textbf{37.2776} & \textbf{37.2515} & \textbf{37.2357} & \textbf{37.2260} \\
    \midrule
    \midrule
    & \multicolumn{5}{c|}{$N = 60$}
        & \multicolumn{5}{c|}{$N = 80$} \\
    Method & 200 & 400 & 600 & 800 & 1000 & 200 & 400 & 600 & 800 & 1000 \\
    \midrule
   \begin{tabular}{@{}ll@{}}
    \pomo (sampling)\\
    \eas \\
    \memento \\
    \end{tabular} &
    
    \begin{tabular}{@{}c@{}}
    37.4598 \\
    37.4569 \\
    \textbf{37.292} 
    \end{tabular} &

    \begin{tabular}{@{}c@{}}
    37.436 \\
    37.3679 \\
    \textbf{37.2607} 
    \end{tabular} &

    \begin{tabular}{@{}c@{}}
    37.4228 \\
    37.3248 \\
    \textbf{37.2305} 
    \end{tabular} &

    \begin{tabular}{@{}c@{}}
    37.4191 \\
    37.2892 \\
    \textbf{37.2171} 
    \end{tabular} &

    \begin{tabular}{@{}c@{}}
    37.4135 \\
    37.2624 \\
    \textbf{37.2092} 
    \end{tabular} &
    
    \begin{tabular}{@{}c@{}}
    37.4588 \\
    37.3588 \\
    \textbf{37.2887} 
    \end{tabular} &

    \begin{tabular}{@{}c@{}}
    37.4335 \\
    37.2804 \\
    \textbf{37.2556} 
    \end{tabular} &

    \begin{tabular}{@{}c@{}}
    37.4198 \\
    \textbf{37.2346} \\
    37.2382 
    \end{tabular} &

    \begin{tabular}{@{}c@{}}
    37.4113 \\
    \textbf{37.2021}\\
    37.2268
    \end{tabular} &

    \begin{tabular}{@{}c@{}}
    37.4050 \\
    \textbf{37.1747} \\
    37.2142 
    \end{tabular} 
    \end{tabular}}
\end{subtable}
\end{table} 

\subsection{Evaluation over larger instances}
In \cref{sec:scaling-rl-construction}, we evaluate \memento and baselines on instances of size 500. For TSP, we use the dataset from \citet{Fu_Qiu_Zha_2021}. For CVRP, we use the dataset from \citet{luo2023neural}. We do not include LEHD in our results since we focus on methods trained with Reinforcement Learning, and LEHD can only be successfully trained with supervised learning at the time of writing. Nevertheless, the good performance achieved by LEHD has motivated us to study the scaling law of \memento as the number of layers in the decoder increases, reported on~\cref{fig:archi-scaling}.

In order to run \compass and \memento(\compass) with the same batch sizes as other methods on those large instances, we reduced the number of starting points used by \compass to ensure that this number multiplied by the number of latent vector sampled at the same time is equal to the number of starting point used by other methods (\pomo, \eas, \memento). In practice, we used a latent vector batch of size 10, and hence divided the number of starting points by 10. This is slightly disadvantaging \compass and \memento(\compass) but enables to respect the constraint of number of parallel batches that can be achieved at once. Note that this slightly impacts the time performance reported since JAX jitting process will not fuse the operations in the same way, additionally, when combined with \memento, this impact the size of the memory (we keep one per starting point to adapt to \pomo, although this is completely agnostic to \memento's method in itself). Since those factors impacts TSP and CVRP performance in different ways, this explains why their relative speed differ, i.e. why \memento(\compass) is slower on TSP but faster on CVRP.

\subsection{Time and memory complexity analysis} \label{subsec:time_complex}

\begin{figure}[ht]
    \centering
    \begin{minipage}{0.45\columnwidth}
        \includegraphics[width=\textwidth]{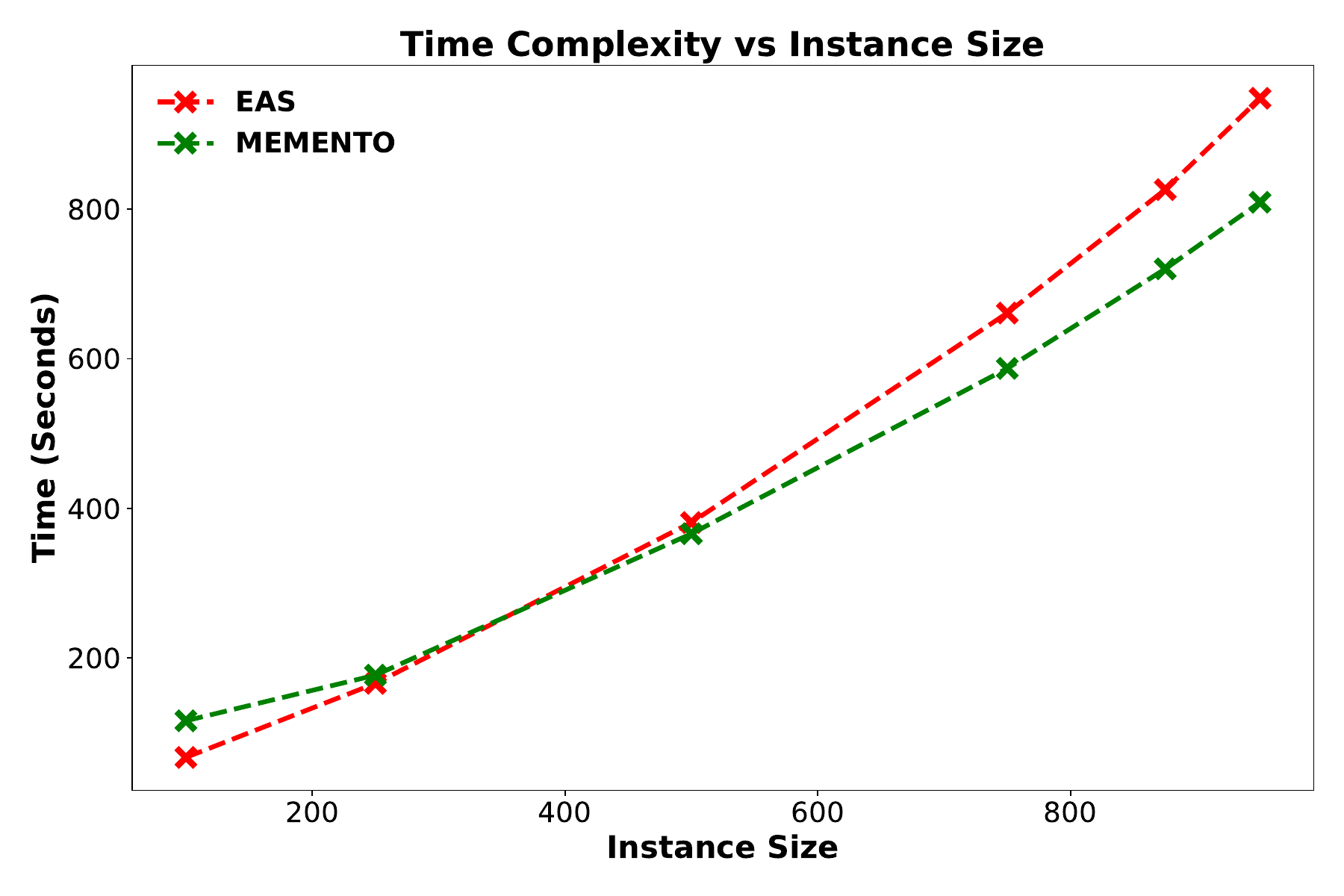}
        \label{fig:instance_scaling}
    \end{minipage}
    \begin{minipage}{0.45\columnwidth}
        \includegraphics[width=\textwidth]{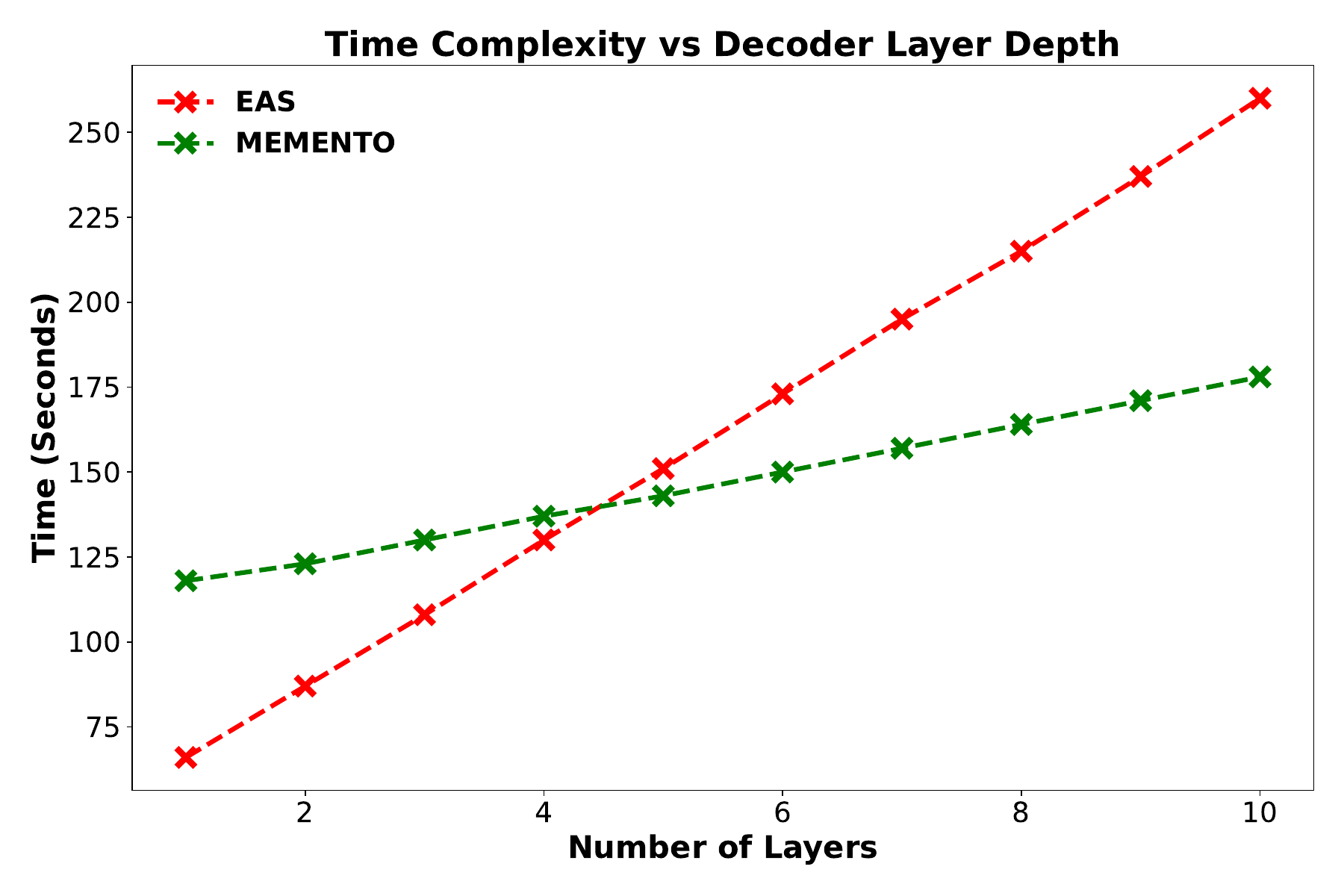}
        \label{fig:archi-scaling}
    \end{minipage}
    \caption{ Time complexity of \memento and \eas for increasing values of decoder size, and instance size. \memento is shown to scale better in time than \eas.}
    \label{fig:size_500_plots}
\end{figure}

\paragraph{Time complexity} To get the time curves reported in~\cref{fig:size_500_plots}, we used CVRP, since it is the environment were \memento was the slowest compared to \eas. We hence expect curves on TSP to be even better for \memento. For the instance size scaling, we use 50 starting points and 100 sequential attempts, and 1 layer in the decoder, and evaluate several instance sizes going from 100 to 1000. For the decoder layer depth scaling, we use CVRP100, 100 starting points and 160 sequential attempts. We then evaluate methods for a depth going from 1 to 10.

\paragraph{Computational and memory complexity analysis}
The use of a memory to compute corrected action logits (i.e., \memento's policy update rule) introduces a computational overhead, whose scaling properties depend on specific design choices and problem parameters.

Consider the TSP100 example with the memory design presented in the paper. The memory contains 100 slots (one per node), each storing 40 entries with 5 float32 values, repeated for each starting point (100). This results in roughly 8 MB of memory data. The memory footprint therefore scales linearly with the instance size (since there is one slot per node), but remains independent of the base policy size. In addition, the MLP module used to process memory entries is small, consisting of two layers with eight hidden units.

Regarding computational cost, the main overhead comes from processing the retrieved entries through the MLP. At each step, 40 entries are retrieved and processed in parallel, so the effective cost is that of 40 independent MLP forward passes. This operation does not depend on the instance size or the policy’s parameter count, but only on the number of retrieved entries. An additional cost arises from indexing the appropriate slice of the memory, which can grow slightly with the number of nodes.

In practice, these overheads make \memento slower than \eas for small instance sizes or small neural networks. However, since its operations scale more favorably with instance size and parameter count (compared to backpropagation), \memento becomes faster for larger problems.

Overall, the computational and memory overhead of \memento is negligible when solving a few instances at a time, typical in practical settings where one receives a new problem instance and a limited compute budget, but it can become a bottleneck when solving large batches of problems in parallel, as often done in research benchmarks.

To provide a more concrete view of the memory-processing cost across instance sizes, we also report the runtime of \pomo in~\cref{tab:runtime-scaling}. The difference between the two indicates the additional cost introduced by \memento. Note that these values may depend on the backend: for example, JAX’s \texttt{jit} compilation can impact relative timings, and alternative implementations (e.g., PyTorch) could yield different ratios.

\begin{table}[h]
\centering
\caption{Average inference time (in seconds) for each algorithm across instance sizes.}
\label{tab:runtime-scaling}
\scalebox{0.8}{
\begin{tabular}{lcccccc}
\toprule
\textbf{Instance Size} & 100 & 250 & 500 & 750 & 875 & 950 \\
\midrule
\pomo     & 57 & 114 & 225 & 360 & 443 & 504 \\
\eas      & 84  & 194 & 446 & 774 & 973 & 1122 \\
\memento  & 148 & 212 & 425 & 684 & 833 & 940 \\
\bottomrule
\end{tabular}}
\end{table}

\section{Training procedure}
\label{appendix:training_process}

This section give a detailed description of how an existing pre-trained model can be augmented with \memento, and how we train \memento to acquire adaptation capacities.

Firstly, we take an existing single-agent model that was trained using the REINFORCE algorithm and reuse it as a base model. In our case, the base model is \pomo. We augment \pomo with a memory module and begin the training procedure which aims to create a policy that is able to use past experiences to make decisions in a multi-shot setting. We initialize the memory module weights with small values such that they barely affect the initial output. Hence, the initial \memento checkpoint maintains the same performance as the pre-trained \pomo checkpoint. 

In the training procedure, the policy is trained to use data stored in the memory to take decisions and solve a problem instance. This is achieved by training the policy in a budgeted multi-shot setting on a problem instance where past experiences are collected and stored in a memory. The memory is organised by nodes such that only information about a specific node is found in a data row that corresponds to that node. The policy retrieves data from the memory at each budget attempt and learns how to uses the data to decide on its next action.

The details of the \memento training procedure are presented in Algorithm~\ref{alg:memento} and can be understood as follows. At each iteration, we sample a batch $\textit{B}$ of instances from a problem distribution $\mathcal{D}$.  Then, for each instance $\rho_i$ where $ i \in  1, \dots , B$, for $K$ budget attempts, we retrieve data from the memory. This is done as follows; given that the current selected node is $a_j$, we only retrieve data $M(a_j)$ associated with node $a_j$ and its starting point. However, the method is agnostic to starting point sampling and would work without it. The features associated with the retrieved action are then normalised. We add remaining budget as an additional feature and process the data by a Multilayer Perceptron (MLP) which outputs a scalar weight for each action. We compute correction logits by averaging the actions based on their respective weights. The correction logits are added to the logits of the base policy. We then rollout the resulting policy $\pi_{\tilde{\theta}}$ on the problem instance (i.e., generate a trajectory which represents a solution to the instance). After every policy rollout attempt, the memory is updated with transitions data such as the action taken, the obtained return, the log-probability of the action and the log-probability of the trajectory. These data are stored in the row that corresponds to the current selected node.

\paragraph{Details about the loss} For each problem instance, we want to optimise for the best return. At each attempt, we apply the rectified linear unit (ReLU) function to the difference between the last return and the best return ever obtained. We use the rectified difference to compute the $\FuncSty{REINFORCE}$ loss at each attempt to avoid having a reward that is too sparse and perform back-propagation through the network parameters of our model (including the encoder, the decoder and the memory networks). The sum of the rectified differences is equal to the best return ever obtained over the budget. As the budget is used, it becomes harder to improve over the previous best, the loss terms hence getting smaller. We found that adding a weight to the terms, with logarithmic increase, helped ensuring that the last terms would not vanish, and thus improved performance. We provide the mathematical formulation below.

Given a problem instance, we unroll $B$ trajectories that we store iteratively in the memory. Each trajectory $\tau_i$ generates a return $R(\tau_i)$. The advantage for each trajectory is defined as $\tilde{R}(\tau_i) = \max(R(\tau_i) - R_{best}, 0)$, where $R_{best}$ is the highest return found so far. The total loss for updating the policy is calculated using the REINFORCE algorithm: $\mathcal{L} = -\sum_{i=1}^{B} \log(1+\epsilon + i) \tilde{R}(\tau_i) \sum_t \log \pi_M(a_t | s_t, M_t)$, where $\pi_M$ is the policy enriched with the memory $M_t$, using the logits $l_M$ defined in eq.1 in the paper. $\epsilon$ is a small number ensuring that the first term is not zero.

To keep the computations tractable, we still compute a loss at each step, estimate the gradient, and average them sequentially until the budget is reached, at which point we take a gradient update step and consider a new batch of instances.

In practice, we also observe that we can improve performance further by adding an optional refining phase where the base model is frozen, and only the memory module is trained, with a reduced learning rate (multiplied by \num{0.1}), for a few hours.

\begin{algorithm}
        \caption{\memento Training}\label{alg:memento}
        \begin{algorithmic}[1]
            \STATE \textbf{Input:} problem distribution $\mathcal{D}$, problem size $N$, memory $M$, batch size $B$, budget $K$, number of training steps $T$, policy $\mathbb{\pi}_{\theta}$ with pre-trained parameters $\theta$.
            \STATE initialize memory network parameters $\phi$ 
            \STATE combine pre-trained policy parameters and memory network parameters $\tilde{\theta} =(\theta, \phi)$
            \FOR{step 1 to $T$}
                \STATE $\rho_i \leftarrow$ Sample($\mathcal{D}$) $\forall i \in 1, \hdots, B$ 
                \FOR{attempt 1 to $K$}
                    \FOR{node 1 to $N$}
                        \STATE $m_{j} \leftarrow $ Retrieve $(M)$ $\forall j \in 1, \hdots, B$ \; \COMMENT{Retrieve data from the memory}
                        \STATE $\tau_{i}^{j} \leftarrow$ Rollout $(\rho_{i}, \pi_{\tilde{\theta}}( \cdot | m_{j} ))$ $\forall i,j \in 1, \hdots, B$
                    \ENDFOR
                    \STATE $m_{j} \leftarrow f(m_{j}, \tau_{i}^{j}) $ \; \COMMENT{Update the memory with transition data}
                    \STATE $R_{i}^{*} \leftarrow \max(R_{i}^{*}, \mathcal{R}(\tau_{i}^{j})) \ \forall i \in 1, \hdots B$ \; \COMMENT{Update best solution found so far}
                    \STATE $\nabla L(\tilde{\theta}) \leftarrow \frac{1}{\textit{B}} \sum_{i \leq B} \FuncSty{REINFORCE}(\FuncSty{ReLU}(\tau_{i}^{j} - R_{i}^{*}))$ \; \COMMENT{Estimate gradient}
                    \ENDFOR
            \STATE $\nabla L(\tilde{\theta}) \leftarrow \frac{1}{K} \sum_{i=1}^{K} \nabla L(\tilde{\theta})$ \; \COMMENT{Accumulate gradients}
            \STATE $\tilde{\theta} \leftarrow \tilde{\theta} - \alpha \nabla L(\tilde{\theta})$ \; \COMMENT{Update parameters}
            \ENDFOR
        \end{algorithmic}
\end{algorithm}

\section{Hyper-parameters}
\label{app:hyperparameters}

We report all the hyper-parameters used during train and inference time. For our method \memento, there is no training hyper-parameters to report for instance sizes 125, 150, and 200 as the model used was trained on instances of size 100. The hyper-parameters used for \memento are reported in~\cref{tab:memento_params}. Since we also trained \pomo and \compass on larger instances, we report hyper-parameters used for \pomo and \compass in~\cref{tab:pomo_params} and~\cref{tab:compass_params}, respectively.  

\begin{table}[ht]
  \centering
    \caption{The hyper-parameters used in \memento}
     \label{tab:memento_params}
\begin{subtable}[t]{\textwidth}
\centering
 \caption{TSP}
  \label{tab:TSP_params}  
    \scalebox{0.7}{
        \begin{tabular}{l | c | c | c | c | c |}
        Phase & Hyper-parameters & TSP100 & TSP(125, 150) & TSP200 & TSP500 \\
        \midrule
        Train time  & budget & 200 & - & - & 200\\
        & instances batch size & 64 & - & - & 32\\ 
        & starting points & 100 & - & - & 30 \\ 
        & gradient accumulation steps & 200 & - &  - & 400 \\ 
        & memory size & 40 & - & - & 80 \\
        & number of layers & 2 & - & - & 2 \\
        & hidden layers & 8 & - & - & 8 \\
        & activation & GELU & - & - & GELU \\
        & learning rate (memory) & 0.004 & - & - & 0.004 \\
        & learning rate (encoder) & 0.0001 & - & - & 0.0001 \\ 
        & learning rate (decoder) & 0.0001 & - & - & 0.0001 \\ 
    \midrule
    
    \begin{tabular}{@{}ll@{}}
      Inference time \\
    \end{tabular} &

    \begin{tabular}{@{}c@{}}
    policy noise \\
    memory size
    \end{tabular} &

        \begin{tabular}{@{}c@{}}
    1 \\
    40 
    \end{tabular} &

        \begin{tabular}{@{}c@{}}
    0.2 \\
    40 
    \end{tabular} &

        \begin{tabular}{@{}c@{}}
    0.1 \\
    40 
    \end{tabular} &

        \begin{tabular}{@{}c@{}}
    0.8 \\
    40 
    \end{tabular} 
    \end{tabular}}
\end{subtable}
\hfill
\begin{subtable}[t]{\textwidth}
 \centering
 \caption{CVRP}
  \label{tab:CVRP_params}
    \scalebox{0.7}{
    \begin{tabular}{l | c | c | c | c | c |}
    Phase & Hyper-parameters & CVRP100 & CVRP(125, 150) & CVRP200 & CVRP500\\
    \midrule
    Train time  & budget & 200 & - & - & 200 \\
    & instances batch size & 64 & - & - & 8\\ 
    & starting points & 100 & - & - & 100 \\ 
    & gradient accumulation steps & 200 & - &  - & 800 \\ 
    & memory size & 40 & - & - & 40 \\
    & number of layers & 2 & - & - & 2 \\
    & hidden units & 8 & - & - & 8 \\
    & activation & GELU & - & - & GELU \\
    & learning rate (memory) & 0.004 & - & - & 0.004 \\
    & learning rate (encoder) & 0.0001 & - & - & 0.0001 \\ 
    & learning rate (decoder) & 0.0001 & - & - & 0.0001 \\ 
    \midrule
    
    \begin{tabular}{@{}ll@{}}
      Inference time \\
    \end{tabular} &

    \begin{tabular}{@{}c@{}}
    policy noise \\
    memory size \\
    \end{tabular} &

    \begin{tabular}{@{}c@{}}
    0.1 \\
    40 \\
    \end{tabular} &

        \begin{tabular}{@{}c@{}}
    0.1 \\
    40 \\
    \end{tabular} &

        \begin{tabular}{@{}c@{}}
    0.1 \\
    40 \\
    \end{tabular} &

        \begin{tabular}{@{}c@{}}
    0.3  \\
    40 \\
    \end{tabular}
    \end{tabular}}
\end{subtable}
\end{table}

\begin{table}[ht]
  \centering
    \caption{The hyper-parameters used in \pomo}
    \label{tab:pomo_params}
    \begin{subtable}[t]{\textwidth}
      \centering
        \scalebox{0.7}{
        \begin{tabular}{l | c | c | c|}
        Phase & Hyper-parameters & TSP500 & CVRP500\\
        \midrule
        Train time  & starting points & 200 & 200 \\ & instances batch size & 64 & 32 \\ & gradient accumulation steps & 1 & 2 \\ 
        \midrule
        
        \begin{tabular}{@{}ll@{}}
          Inference time \\
        \end{tabular} &
    
        \begin{tabular}{@{}c@{}}
        policy noise \\
        sampling batch size  \\
        \end{tabular} &
    
            \begin{tabular}{@{}c@{}}
        1 \\
        8 \\
        \end{tabular} &
    
            \begin{tabular}{@{}c@{}}
        1 \\
        8 \\
        \end{tabular}
        \end{tabular}}
    \end{subtable}
\end{table}

\begin{table}[ht]
  \centering
    \caption{The hyper-parameters used in \compass}
    \label{tab:compass_params}
    \begin{subtable}[t]{\textwidth}
      \centering
        \scalebox{0.7}{
        \begin{tabular}{l | c | c | c|}
        Phase & Hyper-parameters & TSP500 & CVRP500\\
        \midrule
        Train time  & latent space dimension & 16 & 16 \\ & training sample size & 64 & 32 \\ & instances batch size & 8 & 8 \\ & gradient accumulation steps & 8 & 16 \\
        \midrule
        
        \begin{tabular}{@{}ll@{}}
          Inference time \\
        \end{tabular} &
    
        \begin{tabular}{@{}c@{}}
        policy noise \\
        num. \cmaes components  \\
        \cmaes init. sigma  \\
        sampling batch size  \\
        \end{tabular} &
    
            \begin{tabular}{@{}c@{}}
        0.5 \\
        1 \\
        100 \\
        8 \\
        \end{tabular} &
    
            \begin{tabular}{@{}c@{}}
        0.3 \\
        1 \\
        100 \\
        8 \\
        \end{tabular}
        \end{tabular}}
    \end{subtable}
\end{table}

\section{Model checkpoints}
Our experiments focus on two CO routing problems, TSP and CVRP, with methods being trained on two distinct instance sizes: 100 and 500. Whenever possible, we re-use existing checkpoints from the literature; in the remaining cases, we release all our newly trained checkpoints in the repository \textit{anonymised for the review process}.

We evaluate \memento on two CO problems, TSP and CVRP, and compare the performance to that of two main baselines: \pomo~\citep{POMO} and \eas~\citep{hottung2022efficient}. The checkpoints used to evaluate \pomo on TSP and CVRP are the same as the one used in~\citet{poppy} and~\citet{chalumeau2023combinatorial}, and the \eas baseline is executed using the same \pomo checkpoint. These checkpoints were taken in the publicly available repository \url{https://github.com/instadeepai/poppy}. The \pomo checkpoint is used in the initialisation step of \memento (as described in~\cref{appendix:training_process}). To combine \memento and \compass on TSP100, we re-use the \compass checkpoint available at \url{https://github.com/instadeepai/compass} and add the memory processing layers from the \memento checkpoint trained on TSP100. This checkpoint is also released at \textit{anonymised for the review process}.

For larger instances, we compare our \memento method to three baselines: \pomo, \eas and \compass. Since no checkpoint of \pomo and \compass existed, we trained them with the tricks explained in~\cref{experiments}. The process to generate the \memento checkpoint and the \memento(\compass) checkpoint is then exactly the same. All those checkpoints are available, for both TSP and CVRP.

\section{Implementation details}
The code-base in written in JAX~\citep{jax2018}, and is mostly compatible with recent repositories of neural solvers written in JAX, i.e. Poppy and \compass. The problems' implementation are also written in JAX and fully jittable. Those come from the package Jumanji~\citep{jumanji2023github}. CMA-ES implementation to mix \memento and \compass is taken from the research package QDax~\citep{chalumeau2023qdax}. Neural networks, optimizers, and many utilities are implemented using the DeepMind JAX ecosystem~\citep{deepmind2020jax}.

\section{Can MEMENTO discover the REINFORCE update?}
\label{app:math-reinforce}

In~\cref{sec:memento}, we presented the architecture used by the auxiliary model that processes the memory data to derive the new action logits. The intuition behind this architecture choice is that it should be able to learn the REINFORCE update rule. Indeed the REINFORCE loss associated with a new transition is $R \log(\pi_\theta(a))$, such that 
$\frac{\partial R \log(\pi_\theta(a))}{\partial l_a} = R(1 - \pi_\theta(a))$. Therefore, simply having $H_{\theta_M}(f_{a})$ match $R(1 - \pi_\theta(a))$ would recover a REINFORCE-like update. This is feasible, as $R$ and $\log(\pi_\theta(a))$ are included in the features $f_a$. 

We compare the rule learned by \memento to REINFORCE in~\cref{fig:meta-learned-rules} and provide an analysis of how the rule learned by \memento evolve over different budgets in~\cref{app:update_rule}. In~\cref{app:ablation}, we present an ablation study of the features used in the update rule of \memento, showing how much performance can be gained from the use of more information to derive the update rule.

\section{Ablation Study: MEMENTO input features}
\label{app:ablation}

In \cref{sec:memento}, we present \memento, in particular, we present all the inputs that are used by the neural module to derive the new action logits from the memory data. These inputs, or features, are information associated to each past action taken, and that help decide whether those actions should be taken again or not. In \cref{sec:memento-eas}, we compare the rule learned by \memento compared to the policy-gradient update REINFORCE. This comparison is made on the features that both REINFORCE and \memento use: i.e. the action log probability, and the return (or advantage if a baseline is used). Although REINFORCE only uses those two features, \memento uses more, which enables to refine even more the update, and also to adapt it to the budget remaining in the search process. As a recall, in addition to the log probability and return, \memento uses: the log probability of the full trajectory, the log probability of the rest of the trajectory after the action was taken, the budget at the time the action was taken, the action logit suggested by \memento when the action was taken, and the budget currently remaining.

\begin{figure}[ht]
    \centering
    \includegraphics[width=1.\textwidth]{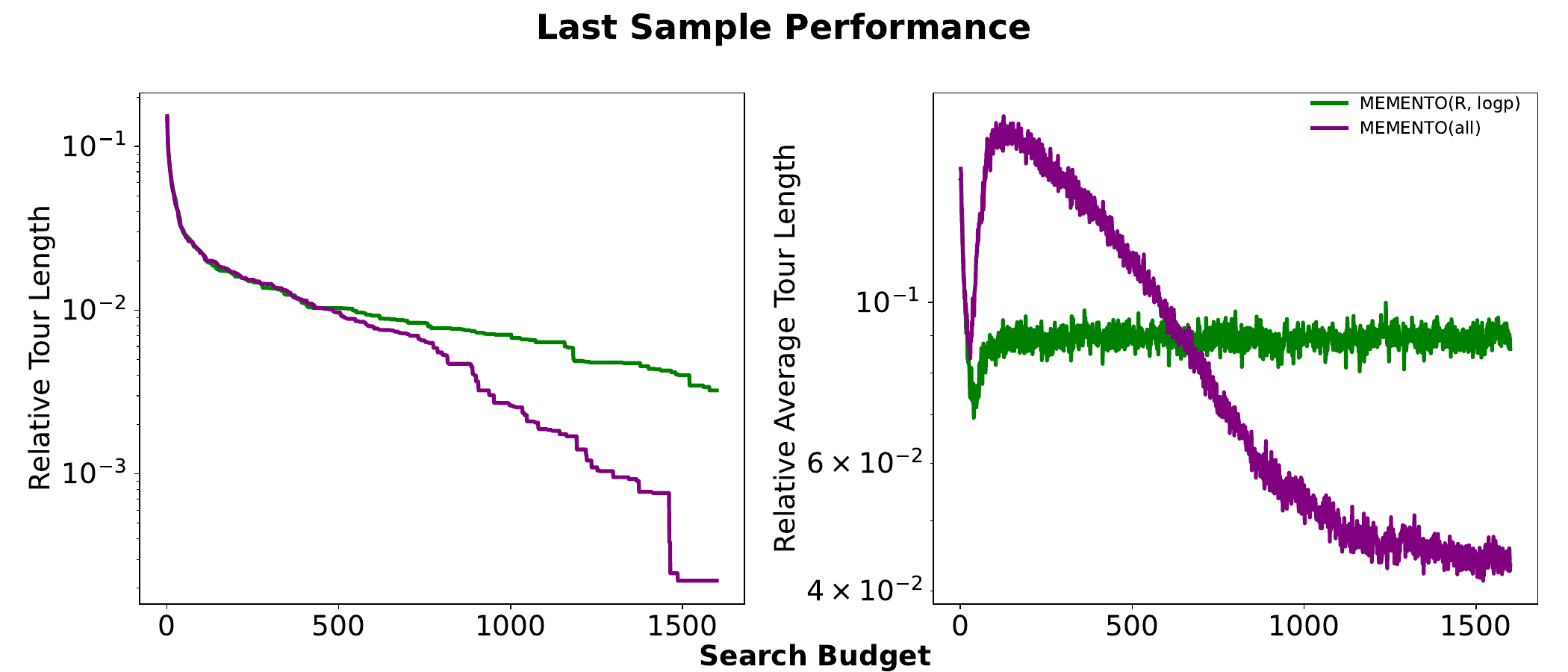}
    \caption{Ablation study of \memento: comparing the impact of memory features that are not available in usual policy gradient estimations methods. The left plot reports the best solution found so far. The right plot shows the performance of the latest solution sampled. The plot illustrates how the additional features enable to achieve a complex exploration strategy, reaching a significantly more efficient adaptation mechanism.}
    \label{fig:ablation-1}
\end{figure}

\begin{figure}[ht]
    \centering
    \includegraphics[width=1.\textwidth]{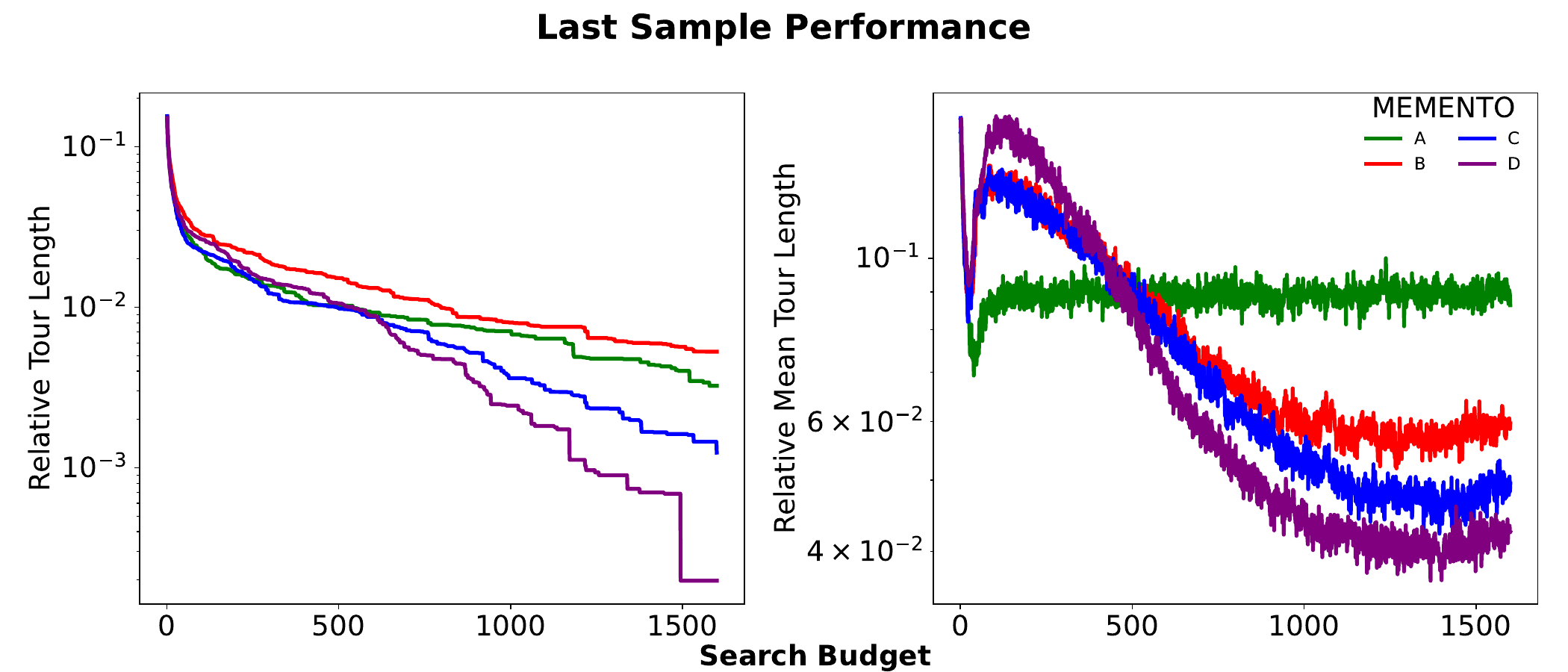}
    \caption{Full ablation study of \memento: comparing the impact of memory features that are not available in usual policy gradient estimations methods. The left plot reports the best solution found so far. The right plot shows the performance of the latest solution sampled. A: return + logp; B: A + remaining budget; C: B + budget when action was taken; D: C + memory logp + trajectory logp + end trajectory logp.}
    \label{fig:ablation-2}
\end{figure}

To validate the impact of all the features used in the memory, we provide an ablation study of those features. To highlight the interest of using all the additional features, we report the performance of \memento using only the return and the log probability, against the performance of \memento with all features, on \cref{fig:ablation-1}. We also report on \cref{fig:ablation-2} a bigger ablation where components are added one after the others.

We can extract two main observations: (i) first, adding the remaining budget completely changes the strategy. We can see on the right panel of \cref{fig:ablation-1} that having access to this additional feature enables \memento to explore much more, and then to focus on high-performing solutions when it gets closer to the end of the budget; (ii) then, \cref{fig:ablation-2} confirms empirically that all features contribute to improving the overall performance.

\section{Evaluation metrics during training phase}

In this section, we provide two plots of \memento's training phase. They show the evolution of performance over time during \memento's training on CVRP100. The left plot reports the evolution of the best tour length obtained during validation over time. The right plot reports the evolution of the improvement delta over time, i.e. the difference between the quality of the best solution generated minus the quality of the first solution generated. This metric shows well how the training phase results in \memento learning an update rule that is able to significantly improve the base policy.

\begin{figure}[ht]
    \centering
    \includegraphics[width=1.\textwidth]{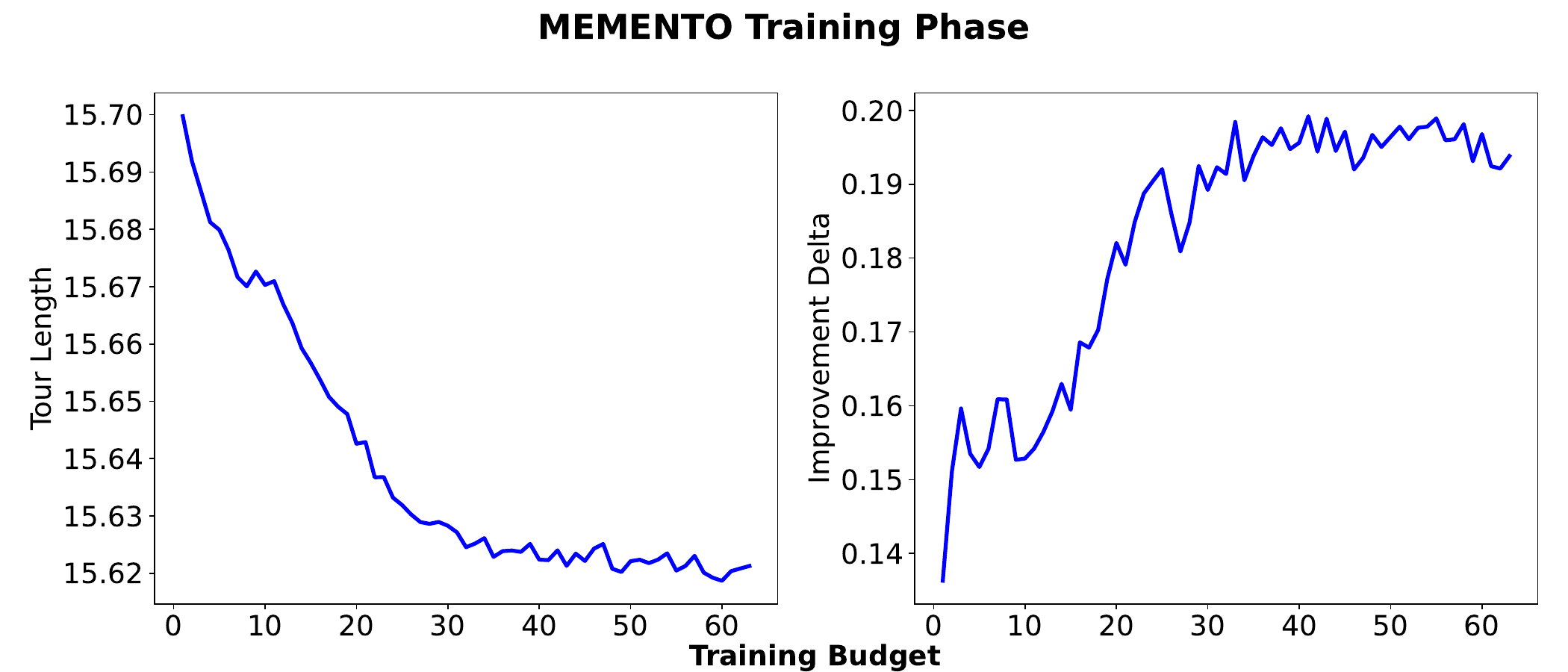}
    \caption{Evaluation metrics during the training phase of \memento.}
    \label{fig:training-plots}
\end{figure}

\section{Memory retrieval mechanism}
\label{app:memory-retrieval}

In this section, we provide additional motivation for the choice of the node retrieval mechanism used in \memento.

Ideally, the retrieval mechanism should retrieve data based on similarity to the current partial solution. But this comes with a computation cost since it requires getting a similarity measure and extracting the k most similar points in the entire memory. We observe that retrieving from the same node is an excellent proxy for similarity, and that the most similar points are very likely to come from the same node. This retrieval strategy hence provides a better trade-off between quality and computation cost, which is why it was selected as the final strategy for \memento.

Nevertheless, \memento comes as a framework, and each practitioner is free to change part of the method to fit each specific problem. One can hence update the retrieval mechanism if desired. Below is an example of an alternative strategy for retrieval that does consider the partial solution constructed. Since the output of the multi-head attention of \pomo’s decoder builds a low-dimensional representation of the partial solution, one can store this vector in the memory, and when building a new solution, retrieve only the k-nearest neighbors of the current partial solution’s representation. One could even apply further dimensionality reduction to reduce the cost of the nearest neighbor search. This approach brings a significant cost increase, even when using state-of-the-art approximated nearest neighbor JAX implementation. This effect gets worse when batching the attempts, or the problem instances. In the problems and settings considered in this paper, our simplified retrieval approach maintains similar results, while significantly improving scaling.

\section{Evolution of MEMENTO learned update over budget}
\label{app:update_rule}
\begin{figure}[ht]
    \centering
    \includegraphics[width=.5\textwidth]{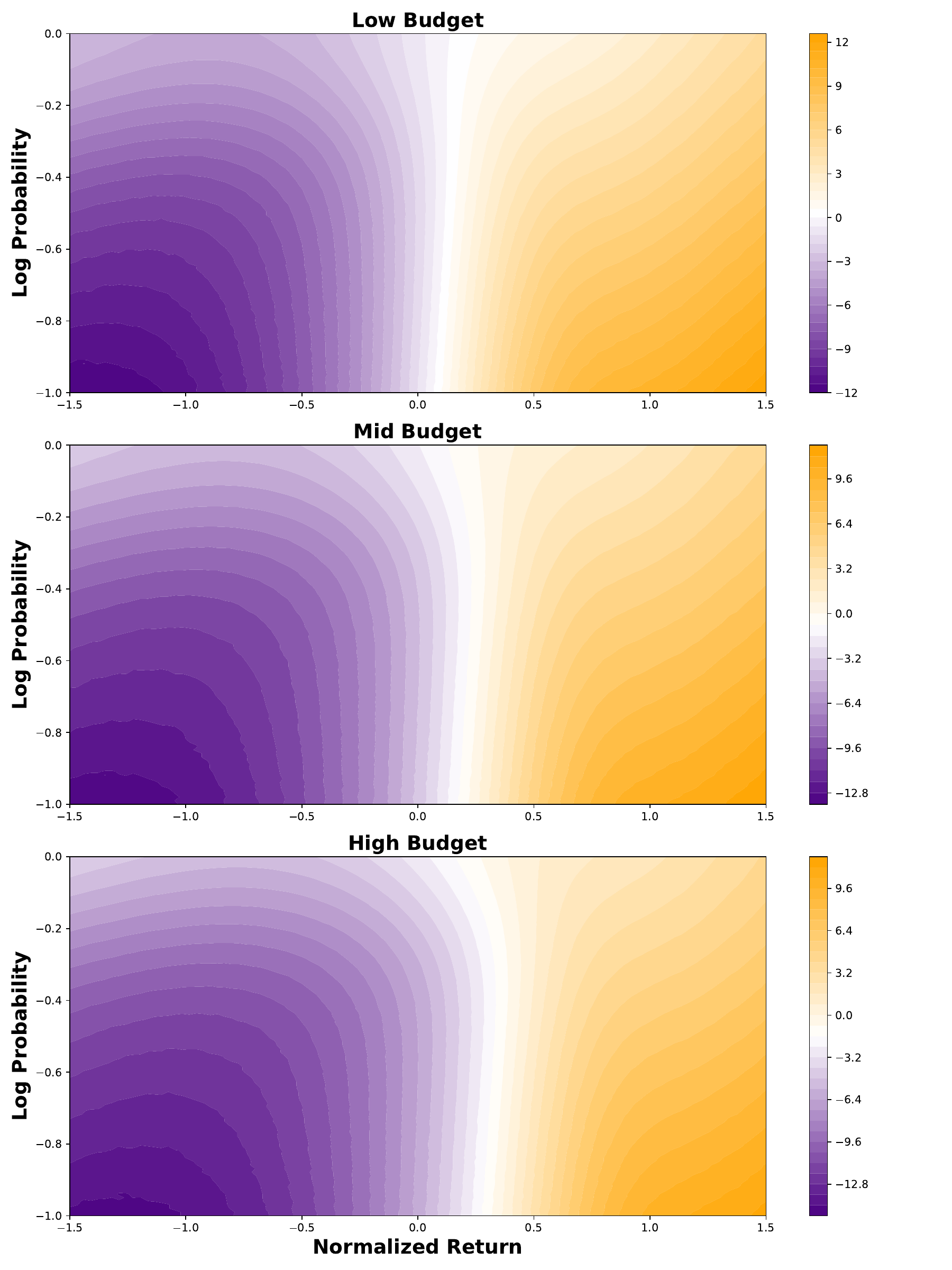}
    \caption{The update rule discovered by MEMENTO across different budget stages (low, mid, high). The learned update rule shifts from broad exploration to sharper exploitation as the budget decreases.}
    \label{fig:update_rule}
\end{figure}

 We further analyse how the update rule learned by the MLP evolves over the course of the budget. Specifically, we compute the MLP’s logit correction values across a grid of inputs by varying the normalised return and the log probability of the action, while fixing all the other inputs. These allows us to isolate the learned update pattern across the return and log probability space. As shown in~\cref{fig:update_rule}, \memento strongly upweights low-probability, high-return actions, consistent with advantage-weighted updates, but exhibits a sharper concentration around high-value transitions. Importantly, we observe a temporal shift: early in the budget, the MLP assigns nonzero corrections even to uncertain or suboptimal transitions promoting broad exploration. As the remaining budget decreases, the corrections become increasingly selective, focusing on high-return, high-confidence actions.

\section{Impact of train-time budget}
\label{app:budget ablation}

The main experiments report results for \memento trained with a budget of 200 sequential attempts. In this analysis, we evaluate variants of \memento trained under alternative training budgets, and evaluate their performance for increasing inference budgets. We report the results in Table~\ref{tab:budget_ablation}.  

Despite being trained with different budgets, all variants of \memento exhibit stable performance across a wide range of inference budgets. This confirms that conditioning on the remaining budget enables flexible behaviour even when the actual inference budget differs from that used in training. Overall, we observe that \memento performs robustly across budget mismatches and that all variants generalise reasonably well beyond their training budget. Notably, the model trained with 200 attempts consistently performs the best which we attribute to a tuning bias: the hyperparameters were selected based on validation performance at budget 200, and were not re-tuned for other settings. 

\begin{table}[ht]
  \centering
    \caption{Performance of \memento on TSP100 across different inference budgets, trained with three budget settings.}
     \label{tab:budget_ablation}
\begin{subtable}[t]{\textwidth}
\centering
   \label{tab:tsp}
    \scalebox{0.5}{
    \begin{tabular}{l | cc | cc | cc | cc | cc | cc | cc | cc | cc |}
     & \multicolumn{18}{c|}{\textbf{Inference Budgets}}\\
     & \multicolumn{2}{c|}{$B=50$} & \multicolumn{2}{c|}{$B=100$} & \multicolumn{2}{c|}{$B=200$} & \multicolumn{2}{c|}{$B=300$} & \multicolumn{2}{c|}{$B=400$} & \multicolumn{2}{c|}{$B=500$} & \multicolumn{2}{c|}{$B=600$} & \multicolumn{2}{c|}{$B=1200$} & \multicolumn{2}{c|}{$B=1600$} \\
     Budget & Obj. & Gap & Obj. & Gap & Obj. & Gap & Obj. & Gap & Obj. & Gap & Obj. & Gap & Obj. & Gap & Obj. & Gap & Obj. & Gap \\ 
    \midrule
    100 & 7.769 & 0.0077\% & 7.767 & 0.0063\% & 7.767 & 0.0058\% & 7.767 & 0.0058\% & 7.767 & 0.0054\% & 7.767 & 0.0056\% & 7.767 & 0.0055\% & 7.767 & 0.005\% & 7.766 & 0.005\%\\
    200 & 7.768 & 0.0074\% & 7.767 & 0.0058\% & 7.766 & 0.005\% & 7.766 & 0.0042\% & 7.766 & 0.0043\% & 7.766 & 0.0042\% & 7.766 & 0.004\% & 7.765 & 0.0035\% & 7.765 & 0.0035\%\\
    300 & 7.768 & 0.0075\% & 7.768 & 0.0066\% & 7.767 & 0.0056\% & 7.767 & 0.0057\% & 7.767 & 0.0057\% & 7.767 & 0.005\% & 7.766 & 0.0043\% & 7.766 & 0.0045\% & 7.766 & 0.0045\%
    \end{tabular}}
\end{subtable}
\end{table}

\end{document}